%% file: acl_latex.tex
\pdfoutput=1

\documentclass[11pt]{article}

\usepackage[final]{acl}

\usepackage{times}
\usepackage{latexsym}
\usepackage{amssymb}
\usepackage{enumitem}

\usepackage{multirow}

\usepackage[T1]{fontenc}

\usepackage[utf8]{inputenc}

\usepackage{microtype}

\usepackage{inconsolata}

\usepackage{graphicx}

\definecolor{iccvblue}{rgb}{0.21,0.49,0.74}
\definecolor{darkgreen}{rgb}{0.0, 0.5, 0.0}
\definecolor{LGray}{gray}{0.97}
\definecolor{ForestGreen}{RGB}{34,139,34}

\usepackage[leftcaption]{sidecap}

 \usepackage{subcaption}
\usepackage{colortbl} 
\usepackage{xcolor}   
\usepackage{booktabs}
\usepackage{xspace}
\usepackage{pifont}
\usepackage{tcolorbox}

\title{A Culturally-diverse Multilingual Multimodal Video Benchmark \& Model}

\author{
    \begin{minipage}[t]{\textwidth}
        \centering
        \normalfont 
         {Bhuiyan Sanjid Shafique}\textsuperscript{1}\thanks{Equal contribution},
         {Ashmal Vayani}\textsuperscript{2}\footnotemark[1],
         {Muhammad Maaz}\textsuperscript{1},
         {Hanoona Abdul Rasheed}\textsuperscript{1} \\
         {Dinura Dissanayake}\textsuperscript{1}, 
         {Mohammed Irfan Kurpath}\textsuperscript{1},
         {Yahya Hmaiti}\textsuperscript{2}, 
         {Go Inoue}\textsuperscript{1},
         {Jean Lahoud}\textsuperscript{1} \\ 
         {Md. Safirur Rashid}\textsuperscript{3}, 
         {Shadid Intisar Quasem}\textsuperscript{3}, 
         {Maheen Fatima}\textsuperscript{4},
         {Franco Vidal}\textsuperscript{2},
         {Mykola Maslych}\textsuperscript{2},
         {Ketan Pravin More}\textsuperscript{1}, 
         {Sanoojan Baliah}\textsuperscript{1},
         {Hasindri Watawana}\textsuperscript{1},
         {Yuhao Li}\textsuperscript{1},
         {Fabian Farestam}\textsuperscript{5},
         {Leon Schaller}\textsuperscript{6}, 
         {Roman Tymtsiv}\textsuperscript{7},
         {Simon Weber}\textsuperscript{6},
         {Hisham Cholakkal}\textsuperscript{1},
         {Ivan Laptev}\textsuperscript{1} \\
         {Shin'ichi Satoh}\textsuperscript{8}, 
         {Michael Felsberg}\textsuperscript{10} 
         {Mubarak Shah}\textsuperscript{2}, 
         {Salman Khan}\textsuperscript{1,9}, 
         {Fahad Shahbaz Khan}\textsuperscript{1,10} \\ [0.5em]
        \small{
            \textsuperscript{1}Mohamed bin Zayed University of Artificial Intelligence, 
            \textsuperscript{2}University of Central Florida,
            \textsuperscript{3}Islamic University of Technology \\
            \textsuperscript{4}Air University,
            \textsuperscript{5}ETH Zurich,
            \textsuperscript{6}	Technische Universität München,
            \textsuperscript{7}Independent Researcher, 
            \textsuperscript{8}National Institute of Informatics \\
            \textsuperscript{9}Australian National University, 
            \textsuperscript{10}Linköping University \\ [0.5em]
            }
        \small{
            \{bhuiyan.shafique, muhammad.maaz, hanoona.bangalath, dinura.dissanayake, mohammedirfan.k, go.inoue, jean.lahoud \\ ketan.more, sanoojan.baliah, hasindri.watawana, hisham.cholakkal, salman.khan, fahad.khan\}@mbzuai.ac.ae \\
            \{ashmal.vayani, yohan.hmaiti, fr543419, mykola.maslych\}@ucf.edu, 
            \{safirurrashid, shadidintisar\}@iut-dhaka.edu \\
            \{leon.schaller, simon1.weber\}@tum.de, 
            \{tymtsiv.roman, lyh88524\}@gmail.com, 
            maheen.fatima@students.au.edu.pk \\ ffarestam@student.ethz.ch,  
            satoh@nii.ac.jp, michael.felsberg@liu.se, ivan.laptev@inria.fr, mubarak.shah@crcv.ucf.edu
            }
    \end{minipage}
}

\begin{document}

\twocolumn[{
\renewcommand\twocolumn[1][]{#1}
\maketitle
\begin{center}
\vspace{5em}
\captionsetup{type=figure}
\includegraphics[width=\textwidth]{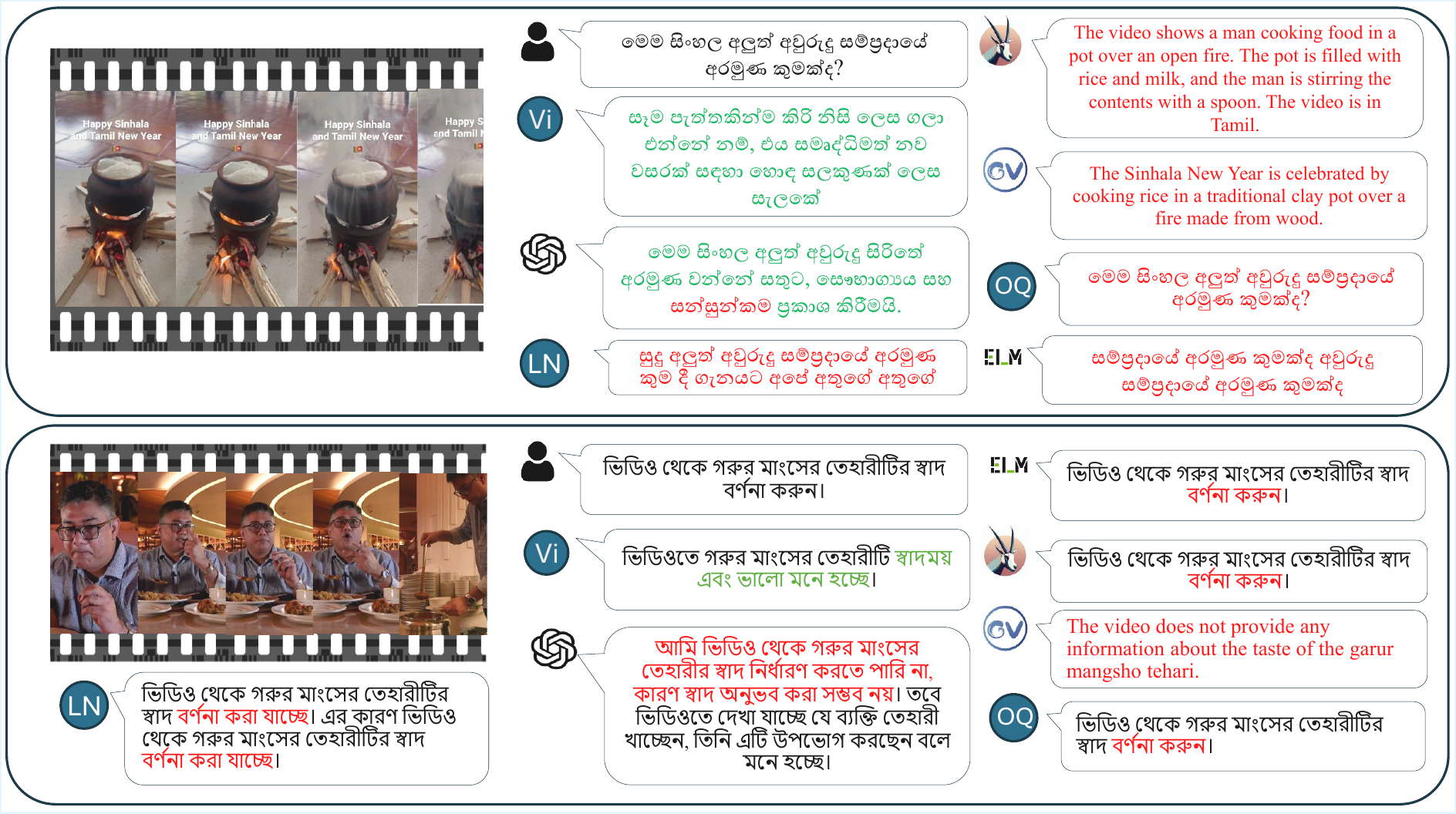}\vspace{0em}
\captionof{figure}{ 
ViMUL-Bench consists of carefully curated videos spanning 14 languages, with 8K manually verified annotations by native experts. It covers 15 diverse domains, incorporating real-world cultural elements such as regional landmarks, local cuisines, and traditional festivals. Additionally, we introduce ViMUL, a simple multilingual baseline designed for general and cultural video comprehension. Qualitative examples (top: Sinhala and bottom: Bengali language) here show that ViMUL performs favorably against recent vidLMMs in cultural inclusivity and overall understanding (errors are highlighted in \textcolor{red}{\textbf{red}} and correct answer in \textcolor{ForestGreen}{\textbf{green}}). ViMUL-Bench covers diverse questions, such as MCQs and short and long visual question answers (VQAs). (\includegraphics[height=1em]{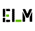}: ViLA, 
    \includegraphics[height=1em]{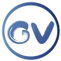}: Video-Chat2, 
    \includegraphics[height=1em]{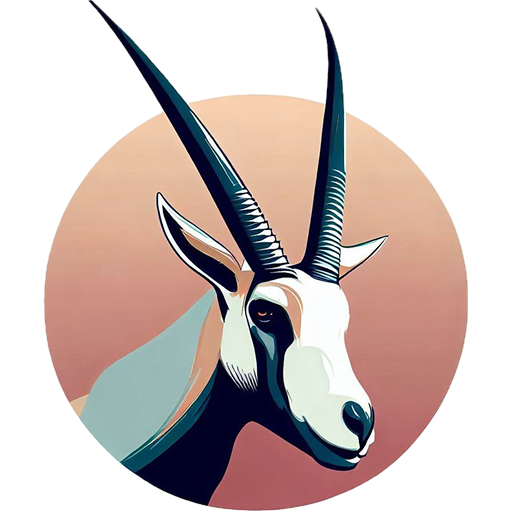}: Video-ChatGPT, 
    \includegraphics[height=1em]{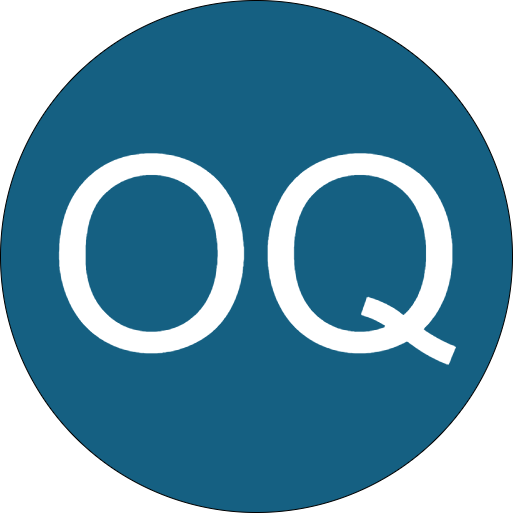}: LLaVA-OneVision-Qwen (OQ), \includegraphics[height=1em]{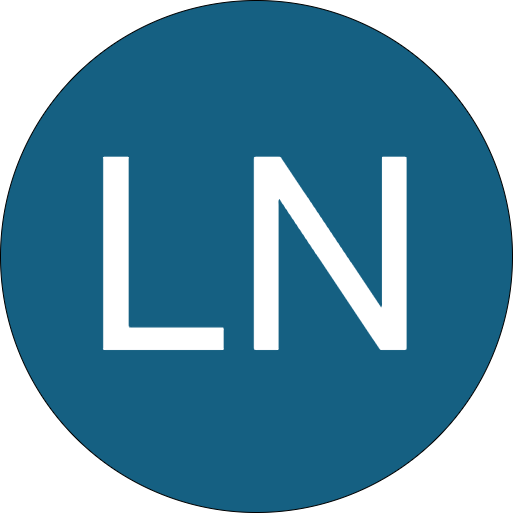}: LLaVA-Next (LN), \includegraphics[height=1em]{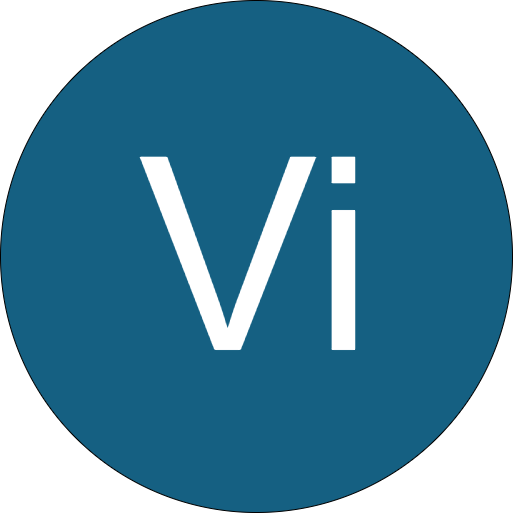}: Our ViMUL).
}
\label{fig:fig_Intro}
\end{center}
}]

\input{sec/0_abstract}    
\input{sec/1_intro}

\input{sec/2_related_works}
\input{sec/3_mint_benchmark}
\input{sec/4_mint_multilingual_lmm}
\input{sec/5_results}
\input{sec/7_conclusion}

\newpage
\input{sec/8_limitations}
\input{sec/6_ethical_consideration}

\bibliography{custom}

\appendix

\label{sec:appendix}
\newpage
\input{sec/X_suppl}

\end{document}

%% file: sec/0_abstract.tex
\begin{figure*}[t!]
    \centering
    \begin{subfigure}[b]{0.48\textwidth}
        \centering
        \includegraphics[width=\textwidth]{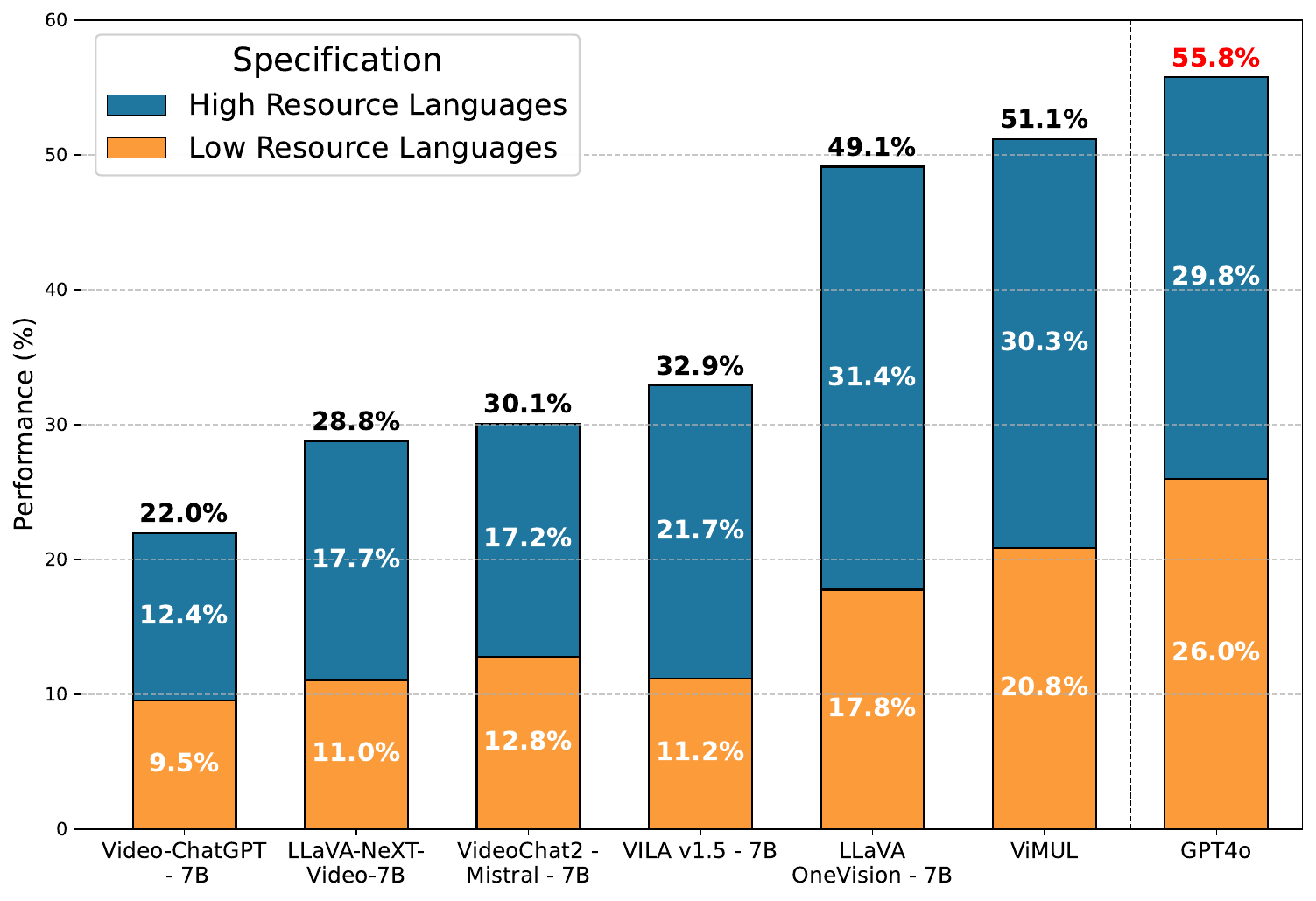}
        \caption{Open vs. Closed-Source LMM performance.}
        \label{fig:open_vs_close}
    \end{subfigure}
    \hfill
    \begin{subfigure}[b]{0.49\textwidth}
        \centering
        \includegraphics[width=\textwidth]{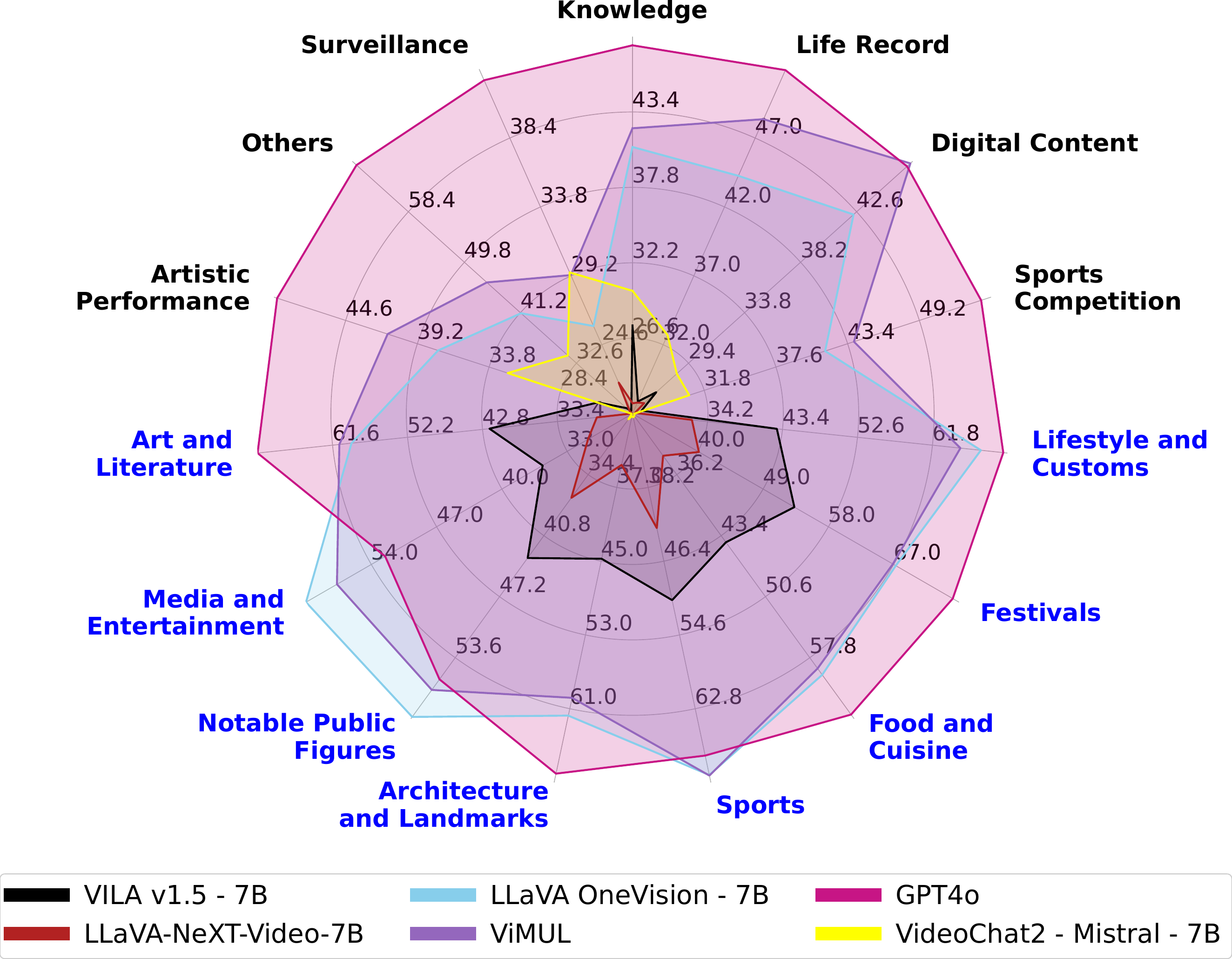}
        \caption{Performance on various general and cultural categories.}
        \label{fig:all_category_plot}
    \end{subfigure}
    \caption{\textbf{Benchmarking video LMMs on the proposed ViMUL-Bench across various languages and cultures}. \textbf{(a)} Performance comparison of open-source versus closed-source models, with a distinction between low-resource and high-resource languages in our ViMUL-Bench. (b) Performance of different video LMMs across 15 diverse categories (both generic and cultural) in our ViMUL-Bench. The categories in \textbf{black} represents generic categories, and categories in \textcolor{blue}{\textbf{blue}} represents the cultural categories.}

    \vspace{-1em}
    \label{fig:category_wise_performance}
\end{figure*}

\begin{abstract}
\footnotetext[1]{Equal contribution}
Large multimodal models (LMMs) have recently gained attention due to their effectiveness to understand and generate descriptions of visual content. Most existing LMMs are in English language. While few recent works explore multilingual image LMMs, to the best of our knowledge, moving beyond the English language for cultural and linguistic inclusivity is yet to be investigated in the context of video LMMs. In pursuit of more inclusive video LMMs, we introduce a multilingual Video LMM benchmark, named ViMUL-Bench, to evaluate Video LMMs across 14 languages, including both low- and high-resource languages: Arabic, Bengali, Chinese, English, French, German, Hindi, Japanese, Russian, Sinhala, Spanish, Swedish, Tamil, and Urdu. Our ViMUL-Bench is designed to rigorously test video LMMs across 15 categories including eight culturally diverse categories, ranging from lifestyles and festivals to foods and rituals and from local landmarks to prominent cultural personalities. ViMUL-Bench comprises both open-ended (short and long-form) and multiple-choice questions spanning various video durations (short, medium, and long) with 8k samples that are manually verified by native language speakers. In addition, we also introduce a machine translated multilingual video training set comprising 1.2 million samples and develop a simple multilingual video LMM, named ViMUL, that is shown to provide a better tradeoff between high-and low-resource languages for video understanding. We hope our ViMUL-Bench and multilingual video LMM along with a large-scale multilingual video training set will help ease future research in developing cultural and linguistic inclusive multilingual video LMMs. Our proposed benchmark, video LMM codebase and training data are available at \url{https://mbzuai-oryx.github.io/ViMUL/}. 

\vspace{-2em}
\end{abstract}

%% file: sec/1_intro.tex
\section{Introduction}
\label{sec:intro}

Large Multimodal Models (LMMs) have achieved remarkable success in vision-and-language tasks, yet their development has predominantly centered on English, overlooking the vast linguistic and cultural diversity of global users \cite{vayani2024all,pfeiffer2021xgqa}. 
This English-centric focus leads to significant performance gaps for other languages, as models often fail to grasp user intent when queries or captions are in low-resource languages. 
Moreover, current models struggle with cultural nuances and region-specific context specifically for low-resource languages \cite{romero2024cvqa, qureshi2025thinking, raza2025responsible}.
For instance, the MaRVL \cite{liu2021visually} and the recent ALM-Bench \cite{vayani2024all} image LMM benchmarks, verified by native speakers to include diverse (including low-resource) languages, reveal dramatic drops in accuracy when state-of-the-art models operate beyond English.
These findings underscore the pressing need for multilingual and multicultural evaluation benchmarks to develop more inclusive next generation of LMMs.

Existing efforts to explore linguistically and culturally diverse LMM benchmarks are limited to \textit{images} \cite{romero2024cvqa, vayani2024all}. To the best of our knowledge, linguistic and cultural diversity are yet to be investigated for \textit{video} LMMs. Video domain poses different challenges as it often depicts complex, culturally rich scenarios-local festivals, foods, rituals, or landmarks, that require understanding both the visual context and the language-specific narration or questions \cite{swetha2025implicitqa, swetha2025timelogic, kim2025safe, sirnam2024x}. Fig. \ref{fig:fig_Intro} illustrates an example where LMMs are asked to describe the taste of the Bengali dish \textit{Beef Tehari}. The models fail to interpret the question in the local language, responding incorrectly and missing linguistic nuances.
While short and long video understanding LMM benchmarks exist in literature, they are typically restricted to only English language.
For instance, Video-MME \cite{fu2024video} focuses on diverse video analysis but in a single language, and MVBench \cite{li2024mvbench} emphasizes temporal reasoning (action sequences, motion) without multilingual considerations.
Other recent efforts like ViLMA \cite{kesen2023vilma} and SEED-Bench \cite{li2023seed,li2023seed2} probe video-language model's abilities in zero-shot temporal grounding and procedural understanding, among other skills, yet none assess cross-lingual or cross-cultural comprehension. 
In short, there is no comprehensive benchmark to evaluate how well video LMMs perform across different languages and cultural contexts (see Tab.~\ref{tab:methods_comparison}).

To bridge this gap, we propose Multilingual Video LMM Benchmark (ViMUL-Bench), the first benchmark for evaluating video LLMs across 14 languages spanning both high-resource and low-resource cases. 
Besides being multilingual, our ViMUL-Bench is designed to test cultural awareness in video LMMs. 
It covers a broad spectrum of culturally diverse categories, including distinct lifestyles, traditional festivals, local cuisines, rituals, regional landmarks, and notable cultural figures.
We formulate a rich evaluation suite with both open-ended questions (requiring descriptive answers in short or long form) and multiple-choice questions (MCQs), curated for videos of varying lengths (short clips, medium snippets, and longer videos) to assess understanding at different temporal scales. 
Crucially, the entire benchmark is verified by native speakers of each language, ensuring that questions and answers accurately capture nuances of tradition, customs, and societal context. 
Additionally, we construct a specialized multilingual video training dataset and train a strong baseline model named ViMUL on it. 
Our experimental analysis reveals that ViMUL provides a better tradeoff between high- and low-resource languages, achieving superior overall performance on multilingual multicultural video question answering, compared to existing open-source video LMMs.  
Our contributions are summarized as follows:
\begin{itemize}
    \item We introduce ViMUL-Bench, a comprehensive benchmark for video LMMs covering 14 languages (including several under-represented ones) and 15 domains, including real-world cultural aspects. To our knowledge, this is the first effort to enable rigorous testing of video LMMs across a wide linguistic and cultural spectrum, emphasizing both cross-lingual and cultural comprehension (see Tab.~\ref{tab:methods_comparison}).
    \item The ViMUL-Bench offers 8K manually verified diverse samples for comprehensive spatio-temporal evaluation and includes both open-ended and multiple-choice QAs. It also offers diversity in terms of video length (short, medium, and long videos). In addition, we provide a multilingual training dataset with 1.2M samples translated from available video datasets.
    \item We propose ViMUL, a multilingual video LLM fine-tuned on our multilingual training set. ViMUL establishes a strong baseline on ViMUL-Bench, providing a better overall tradeoff compared to existing open-source video LMMs for multilingual video understanding (see Fig.~\ref{fig:category_wise_performance}). 
\end{itemize}

%% file: sec/2_related_works.tex
\section{Related Works}
\label{sec:related_works}

\noindent\textbf{Multilingual Multicultural Datasets:}
Early vision-language benchmarks were predominantly English-centric, with limited coverage of other languages or cultures \cite{romero2024cvqa}.
Recent efforts have sought to bridge this gap by extending multimodal tasks to multiple languages.
For example, xGQA expanded the GQA visual question answering dataset to seven diverse languages via translation \cite{pfeiffer2021xgqa}.
However, such translation-based approaches often reuse the same generic/biased image distributions and thus fail to capture cultural nuances \cite{romero2024cvqa}.
To introduce culture-specific content, \cite{liu2021visually} proposed a Multicultural Reasoning dataset (MaRVL).
While MaRVL incorporates diverse concepts, its scope is limited (five languages and binary true/false reasoning).
Similarly, other contemporary benchmarks include M3Exam \cite{zhang2023m3exam}, MMMB \cite{sun2024parrot}, MMBench \cite{liu2024mmbench}, M4U \cite{wang2024m4u}, and Exams-V \cite{das2024exams} which introduce multilingual evaluation samples for image understanding. 
However, their scope is generally limited to a few languages and offers a narrow cultural scope.
More recent benchmarks extend the scope further, e.g., CVQA \cite{romero2024cvqa} introduces 10K VQA examples grounded in 31 languages (13 scripts), using culturally relevant images and human-curated questions.
The recent All Languages Matter (ALM-Bench) benchmark \cite{vayani2024all} spans 100 languages with images drawn from 13 distinct cultural aspects. 
It is the largest multicultural multimodal evaluation benchmark, designed to test LMMs on diverse imagery and low-resource languages. 
However, these benchmarks remain limited to the image-domain and do not address multilingual and multicultural aspects unique to videos.

\input{tables/benchmark_comparison}



\begin{figure*}[!t]
  \centering
  \includegraphics[width=0.9\linewidth]{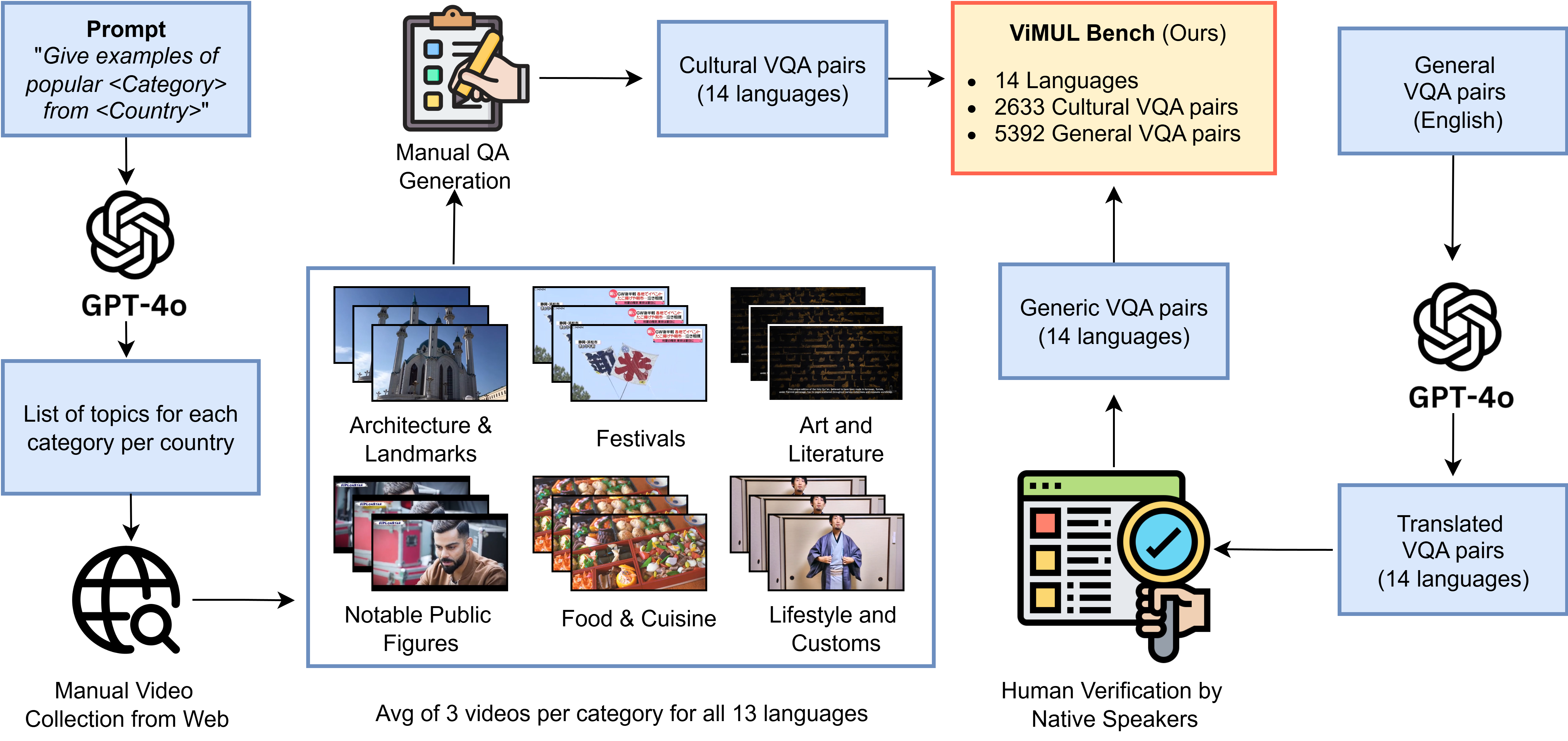}
  \caption{\small \textbf{Data collection and verification pipeline.} Our benchmark consists of both cultural-specific video content curated from scratch \textit{(left)} and generic Video-QA pairs sourced from existing video LMM benchmarks. Cultural videos are scrapped using a \textit{(country, language, sub-topic)} triplet and manually filtered for relevance and private information. With the help of native speakers, we create QA pairs for each language from scratch (except English), with cultural QA pairs translated into English using GPT-4o. Our ViMUL-Bench has diverse question types and features approximately 8K QA pairs in 14 languages.}
  \label{fig:mint_annotation_pipeline_dia}
\end{figure*}

\noindent\textbf{Video LMM Benchmarks:} 
Extending multimodal evaluation to video introduces additional challenges due to temporal dynamics \cite{swetha2021unsupervised, swetha2023preserving}.
A number of video-language benchmarks have emerged to assess LMMs on video understanding, though they focus on general monolingual capabilities.
Video-MME \cite{fu2024video} is introduced as the first comprehensive evaluation of LMMs on video analysis, covering a broad spectrum of video domains.
It offers a full-spectrum evaluation, covering questions from perception to reasoning, and variable videos lengths ($\sim$10 secs to $\sim$1 hour).
Another effort, MVBench \cite{li2024mvbench} concentrates on temporal reasoning skills and proposes an automatic pipeline to generate a large multiple-choice QA benchmark by leveraging existing video datasets and GPT-based annotators.
Beyond general benchmarks, ViLMA \cite{kesen2023vilma} takes a fine-grained approach as it uses carefully designed counterfactual video pairs to probe a model’s temporal grounding and linguistic understanding in a zero-shot setting.
SEED-Bench is another comprehensive multimodal benchmark that includes some video-based questions; however, those are largely confined to temporal or procedural understanding tasks \cite{li2023seed,li2023seed2}.
These benchmarks have a significantly advanced evaluation for video-based multimodal reasoning and generation, however they lack multilingual and culturally aware components. 
Current video benchmarks evaluate models primarily on English video-question pairs and generic content, without testing performance on non-English dialogues, region-specific contexts, or culturally diverse visual narratives. 
The absence of multilingual and multicultural evaluation for video LMMs forms a key motivation for ViMUL-Bench, which aims to fill that gap by assessing video understanding across diverse languages and cultural settings.

%% file: tables/benchmark_comparison.tex
\begin{table}[!t] 

\centering
    \setlength{\tabcolsep}{2pt}
    \renewcommand{\arraystretch}{1.5}
    \resizebox{\columnwidth}{!}{%

    {\Huge
    \begin{tabular}{lcccccccccccccc}
    
        \toprule
        \textbf{Benchmark} & \textbf{Multi-} & \textbf{Total} & \textbf{Total}  & \textbf{Question} & \textbf{Total} & \textbf{Annotation} & \textbf{Cultural}  \\
        & \textbf{lingual} & \textbf{\hspace{0.5cm}Domains\hspace{0.5cm}} & \textbf{Samples}  & \textbf{Types} & \textbf{Videos} & \textbf{Type} & \textbf{Content}
        \\
        \midrule

        \rowcolor{LGray} ActivityNet-QA &  &  &  &  &  &  &  \\

        \rowcolor{LGray} \cite{caba2015activitynet} & \textcolor{red}{\ding{55}} (1)  & - & 2378  & OE & 5800 & Human & \textcolor{red}{\ding{55}}  \\


        CVRR-ES \cite{khattak2024good} & \textcolor{red}{\ding{55}} (1)  & - & 2400  & OE & 224 & Auto & \textcolor{red}{\ding{55}} \\

        \rowcolor{LGray} MoVQA \cite{zhang2023movqa} & \textcolor{red}{\ding{55}} (1)  & 6 & 21,953  & OE & - & Human & \textcolor{red}{\ding{55}}  \\

        MovieQA \cite{tapaswi2016movieqa} & \textcolor{red}{\ding{55}} (1)  & - & 6462  & MCQ & 6771 & Human & \textcolor{red}{\ding{55}}  \\

        \rowcolor{LGray} MSVD-QA \cite{xu2017video} & \textcolor{red}{\ding{55}} (1)  & 5 & 50,505  & OE & 1970 & Auto & \textcolor{red}{\ding{55}} \\

        MVBench \cite{li2024mvbench} & \textcolor{red}{\ding{55}} (1)  & - & 4000  & MCQ & 3507 & Auto & \textcolor{red}{\ding{55}} \\

        \rowcolor{LGray} Perception Test \cite{patraucean2023perception} & \textcolor{red}{\ding{55}} (1)  & - & 44,000  & MCQ+OE & 11,600 & \hspace{0.3cm}Auto+Human & \textcolor{red}{\ding{55}} \\

        TVQA \cite{lei2018tvqa} & \textcolor{red}{\ding{55}} (1)  & - & 15,2545  & MCQ+OE & 21,793 & Human & \textcolor{red}{\ding{55}} \\
        

        \rowcolor{LGray} Video-MME \cite{liu2021visually} & \textcolor{red}{\ding{55}} (1)  & 6 & 2700  & MCQ & 900 & Human & \textcolor{red}{\ding{55}} \\

        VCG Diverse \cite{Maaz2024VideoGPT+} & \textcolor{red}{\ding{55}} (1)  & 18 & 4354  & SVQA,LVQA & 877 & Auto & \textcolor{red}{\ding{55}} \\
        
        \rowcolor{violet!10}\textbf{Ours} & \textcolor{darkgreen}{\checkmark} (14)  & 15 & 8,025  & MCQ,SVQA,LVQA  & 879 & Auto+Human & \textcolor{darkgreen}{\checkmark} \\



        \bottomrule
    \end{tabular}
    }
    }
    \vspace{-0.5em}
    \caption{Comparison of video LMM benchmarks emphasizing multilingual and cultural understanding. \textit{Domains} represent the aspects covered by each dataset for different languages. \textit{Annotation Type} is categorized as follows: Human - Questions were created in the local language. Human+Auto - Questions were generated or translated using GPT-4/Google API and later validated by human experts. Auto: Questions were generated or translated automatically without human validation. `-' indicates that information is not available.}

    \vspace{-1.5em}
    \label{tab:methods_comparison}
\end{table}

%% file: sec/3_mint_benchmark.tex
\section{ViMUL-Bench}
\label{sec:mint_benchmark}

ViMUL-Bench is a comprehensive multilingual benchmark, designed to evaluate both general and culturally-specific aspects of video comprehension in video LMMs. It captures rich cultural nuances through a diverse set of question-answer (QA) pairs, including multiple-choice and open-ended (short and long) formats \cite{raza2025vldbench, campos2025gaea}. The benchmark spans 15 diverse categories, categorized into general and cultural topics, across 14 languages: Arabic, Bengali, Chinese, English, French, German, Hindi, Japanese, Russian, Sinhala, Spanish, Swedish, Tamil, and Urdu. 
We build upon the recent PALO \cite{Maaz_2025_WACV}, incorporating its 10 languages while adding Swedish, German, Tamil, and Sinhala to ensure typological diversity and to enhance the representation of low-resource languages such as Tamil, Urdu, and Sinhala as defined by Ethnologue \cite{campbell2008ethnologue} and Glottolog \cite{glottolog2022} database.

The \textit{generic} category includes seven domains: Artistic Performance, Digital Content, Knowledge, Life Record, Sports Competition, Surveillance, and Others. The \textit{cultural} categorization is inspired from recent image LMM benchmarks \cite{vayani2024all, romero2024cvqa, marino2019ok}, where the corresponding videos for each domain and language are manually scrapped with their annotations manually verified by a native speaker. It spans eight diverse domains, including Lifestyle \& Customs, Festivals, Food \& Cuisine, Sports, Architecture \& Landmarks, Notable Public Figures, Media \& Entertainment, and Art \& Literature. Our ViMUL-Bench has been meticulously curated and verified by native experts for the 13 languages to ensure high-quality question answer (QA) pairs that accurately capture the nuances of all 15 domains. It comprises 8,025 diverse questions in total across 14 languages, spanning both generic and cultural categories to comprehensively evaluate multilingual and cross-cultural video understanding.


\begin{figure*}[!t]
  \centering
  \includegraphics[width=1\linewidth]{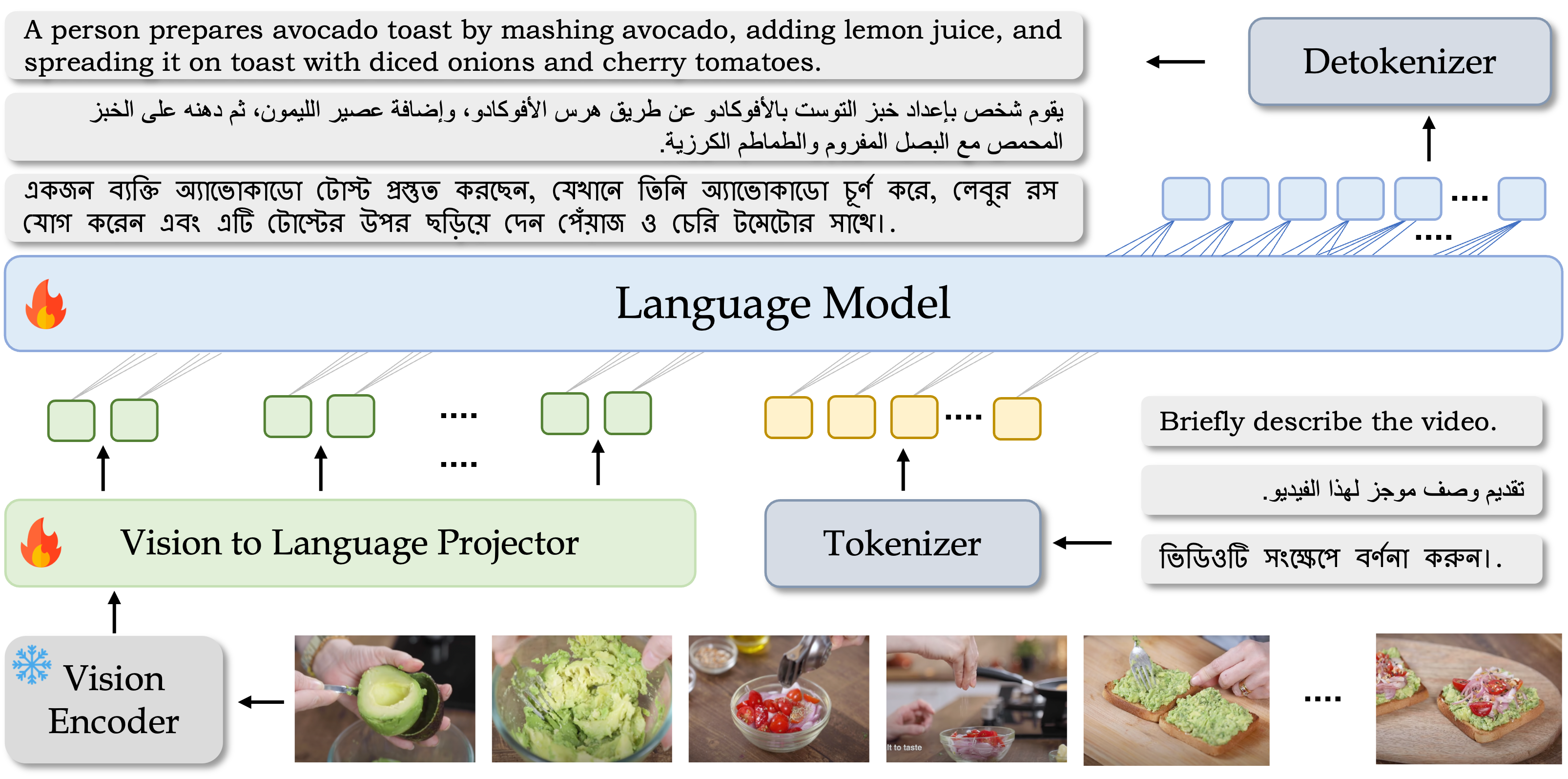}
  \caption{\small \textbf{Overview of ViMUL}. 
    ViMUL is designed to comprehend and generate content in 14 different languages: Arabic, Bengali, Chinese, English, French, German, Hindi, Japanese, Russian, Sinhala, Spanish, Swedish, Tamil, and Urdu, covering at least two-thirds of the global population. The model employs a vision encoder to process video frames, followed by a vision-to-language projector and an LLM. The projected features are then concatenated with the user query and fed into the LLM to generate a response. 
    (\includegraphics[height=1em]{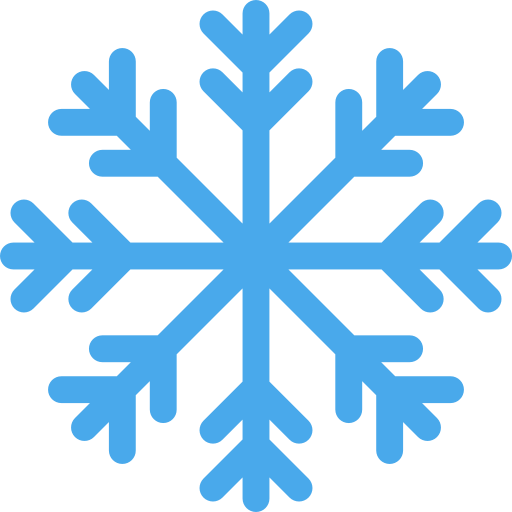}: frozen, \includegraphics[height=1em]{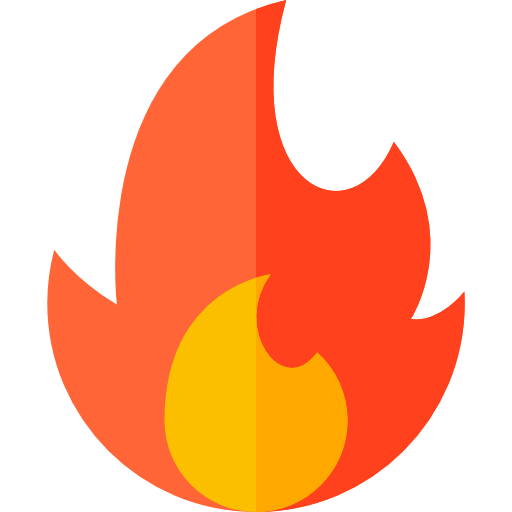}: trained)}
  \label{fig:mint_lmm_block_dia}
\end{figure*}

\subsection{Data Collection and Annotation}
\noindent\textbf{Generic VQA Curation.} As discussed above, our ViMUL-Bench encompasses both generic and cultural categories to evaluate multilingual video understanding across 14 languages. For the generic category, we carefully curate English-only subsets from VCG-Diverse \cite{Maaz2024VideoGPT+}, CVRR-ES \cite{khattak2024good}, MVBench \cite{li2024mvbench}, and VideoMME \cite{fu2024video}. Among these, VCG-Diverse and CVRR-ES follow an open-ended question format, while MVBench and VideoMME use multiple-choice questions (MCQs). These English samples are then translated into 13 additional languages using GPT-4o, followed by manual verification and refinement by native speakers of each language. Language experts were explicitly instructed to correct poor translations, ensuring accuracy by rephrasing or rewriting question-answer pairs when necessary. This process resulted in 5,392 QA pairs from 542 videos, in the generic category (see Fig. \ref{fig:mint_annotation_pipeline_dia}).

\noindent\textbf{Cultural Video Curation.}
To curate diverse cultural QA pairs, we collect open-licensed videos and their corresponding metadata from the internet, focusing on specific cultural aspects of each language across three durations: short (0-4 mins), medium (4-15 mins), and long (15+ mins). To accurately capture the cultural nuances of each language, we follow the approach outlined in \cite{vayani2024all, romero2024cvqa} and link each language to a country based on the World Values Survey \cite{haerpfer2022world} to ensure coverage of cultural and ritual diversity. Additionally, we generate a list of topics for each category, forming a triplet (language-country-topic), using GPT-4o, and then search for relevant content. For example, querying ``Popular sports in {United States}" yields responses such as ``baseball, soccer, golf, and ice hockey." We manually extract videos in the native language to ensure linguistic accuracy. Each domain undergoes several filtration steps, such as removing low-resolution, noisy, or unclear videos, to guarantee data quality. To maintain both high-quality and cultural relevance, we enlist expert native speakers of each language to manually verify the quality and cultural diversity of videos. Any content lacking cultural relevance is removed from the dataset. Fig. \ref{fig:mint_annotation_pipeline_dia} shows our data collection and verification pipeline.

\noindent\textbf{Cultural QA Generation.} To generate high-quality video-QA pairs for the cultural section of ViMUL-Bench, the question-answer (QA) pairs are curated via native experts based on the provided videos and their metadata. Notably, videos and their corresponding QAs are not shared across languages in the cultural set. For each video, we generate multiple-choice questions (MCQs) and one open-ended question in the native language. To reduce randomness in MCQ generation, we ensure that each question can also be answered when rephrased as an open-ended question. Further, the native experts are tasked with creating an English version of the question-answer pair. We instruct the experts to focus on cultural concepts depicted in each video and to generate questions that require a visual understanding of the video, while avoiding the perpetuation of bias and stereotypes. Following this process, 2,633 QA pairs were curated, spanning 337 videos for the cultural category.

%% file: sec/4_mint_multilingual_lmm.tex
\section{ViMUL: Multilingual Video LMM}
\label{sec:mint_lmm}

\begin{figure}[!t]
    \centering
    \includegraphics[width=\columnwidth]{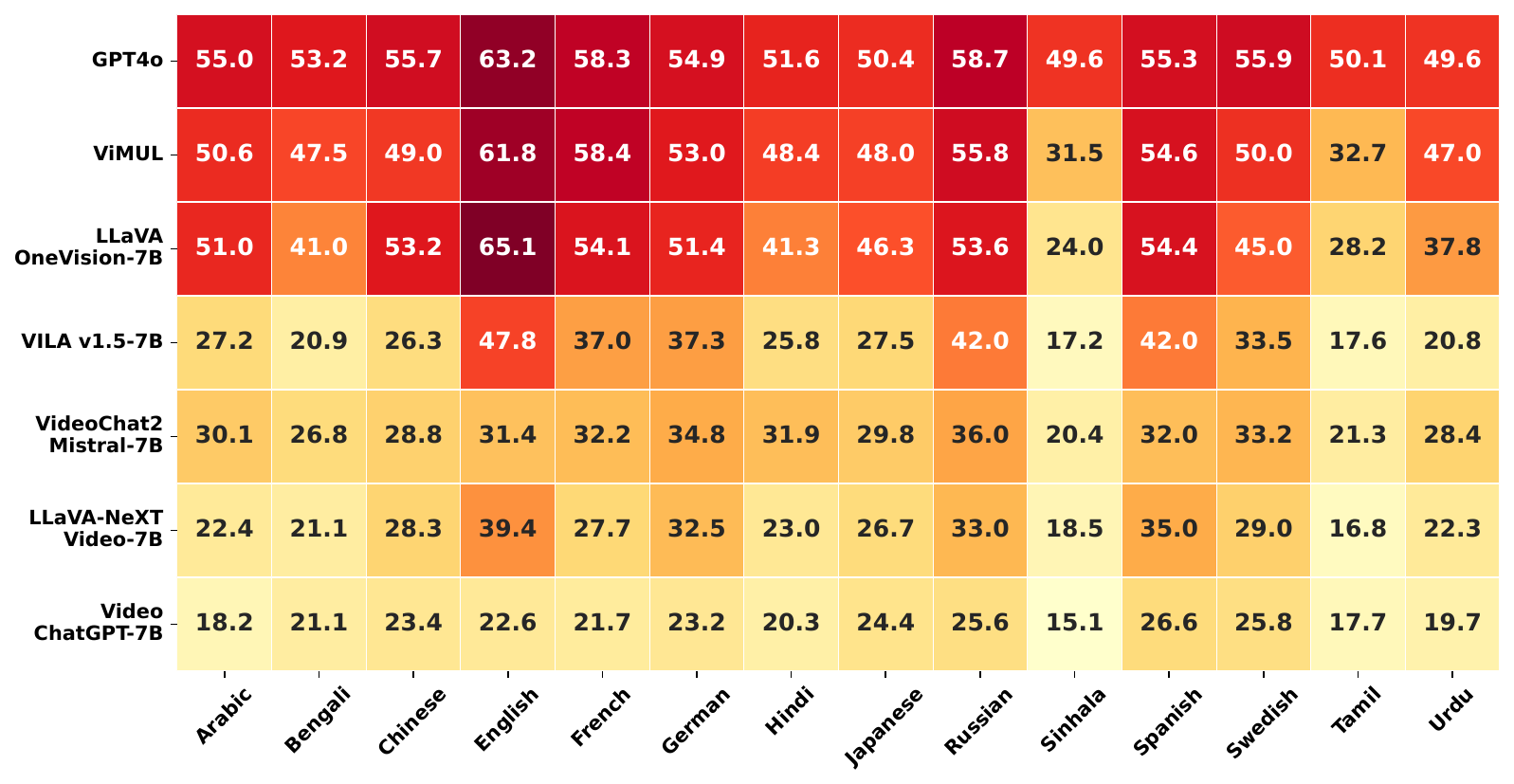}
    \vspace{-2em}
    \caption{\small \textbf{Performance comparison of video LMMs across 14 languages on ViMUL-Bench}.
    Average accuracy is reported across all question types for each language. Each box represents a model’s accuracy for a specific language, with darker shades indicating higher accuracy. The results show that the closed-source model, GPT-4o, generally outperforms its open-source counterparts.
    In contrast to high-resource languages, methods struggle on low-resource languages (e.g., Sinhala, Urdu, Tamil). Among open-source models, our ViMUL provides a better tradeoff between high and low-resource languages, achieving an overall gain of 2\% over LLaVA-OneVision.}
    \vspace{-1.5em}
    \label{fig:heatmap_results}
\end{figure}

In addition to the ViMUL-Bench, we develop a multilingual video LMM, named ViMUL, by constructing a multilingual video training set. ViMUL is built to understand and generate content in 14 diverse languages, covering an audience that represents at least two-thirds of the world's population.

\noindent
\textbf{Overall Architecture:}  
The architecture of ViMUL is derived from LLaVA-OneVision~\cite{llava_onevision}, which seamlessly integrates a vision encoder, a vision-to-language projector, and a language model. The video frames are encoded using a vision encoder, projected into the language model's embedding space using a two-layer MLP projector, concatenated with the text embeddings, and passed to the language model to generate the response (see Fig.~\ref{fig:mint_lmm_block_dia}). We discuss the Video Sampling strategy in Sec. \ref{mutlilingual_videolmm} (suppl. material).

\subsection{Multilingual Instruction Tuning Dataset}
One of the contributions of this work is the development of a comprehensive multilingual video-language instruction-tuning dataset. Current video instruction-tuning datasets exist only in English and do not focus on other languages. However, recent advances in large language models (LLMs) have demonstrated impressive performance in multilingual tasks. We leverage these advancements and use GPT-4o-mini~\cite{2024GPT4o} to translate the video instruction-tuning dataset from English to 13 additional languages, thereby creating a multilingual dataset that broadens the linguistic scope and applicability of the model.
Our dataset is sourced from two primary sources: Video-Instruct100K~\cite{Maaz2023VideoChatGPT} and LLaVA-Video-178K~\cite{llava_178K}. 
Video-Instruct100K is a video instruction-tuning dataset containing 100K samples generated using a semi-automatic annotation pipeline. 
The QA pairs are open-ended, including both short and long question-answer formats. We use the human-verified version of Video-Instruct100K released by VideoGPT+~\cite{Maaz2024VideoGPT+}, which consists of 25,803 samples. 
The videos in Video-Instruct100K are sourced from the ActivityNet dataset~\cite{caba2015activitynet}. Finally, our pipeline results in a total of 1,238,102 samples across all languages. We also discuss dataset-wise, per-language QA-pair distribution, and common translation issues in Sec. \ref{mutlilingual_videolmm} (suppl. material).

To evaluate the translation quality generated by GPT-4o, we perform a \textit{cycle consistency check} \cite{hu2011linear}. We randomly sample 10,000 QA pairs across 10 languages, translate them back to English using Qwen-3 \cite{yang2025qwen3}, and compare the results with our original English subset using GPT-4o as the judge. Fig.~\ref{fig:cycle_consistency} (suppl. material) shows the per-language translation scores. The performance ranges from 95.3\% in French to 84.4\% in Bengali, demonstrating the quality of our multilingual data.

\begin{figure*}[!t]
  \centering
    \includegraphics[width=1.0\linewidth]{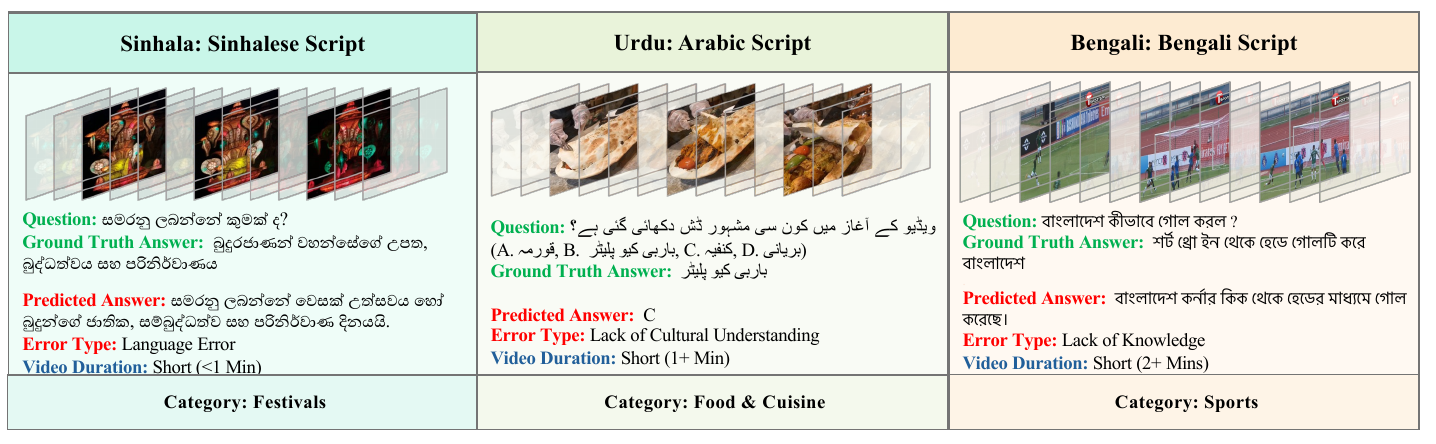}\vspace{-0.8em}
    \caption{\small We present qualitative examples of failure cases of GPT-4o’s across different language scripts and categories, specifying the corresponding error types. For instance, in a Sinhala-language question asking about the event being celebrated in the video, the model correctly identifies the cultural significance-celebrating the birth, enlightenment, and passing of Lord Buddha, but fails to respond with grammatically correct Sinhala, highlighting a language proficiency error. Results on success cases are shown in Fig. \ref{fig:qual_success_cases} (suppl. material).}
    \label{fig:qual_error_types}

    \vspace{-1.5em}
\end{figure*}

%% file: sec/5_results.tex
\section{Results and Discussions}

As discussed earlier, ViMUL-Bench comprises two question types: multiple-choice and open-ended, including both short and long question-answer formats, different prompts are employed for each question type. For multiple-choice questions, we provide the visual context and textual query to the LMMs, instructing them to select the best option, which is then directly compared to the ground truth. Performance is measured using \textit{accuracy}, following established multiple-choice VQA benchmarks \cite{romero2024cvqa, bang2023multitask, zhu2016visual7w}. For open-ended questions, we use the open-source multilingual LLM, Phi-4-14B \cite{abdin2024phi} as a judge, ensuring consistency and reproducibility, unlike GPT-based models \cite{shen2023large, stureborg2024large}, which are costly and inconsistent due to version updates. Performance is evaluated using \textit{correctness} criteria, which measure how closely the model's output matches the ground truth (see Sec. \ref{promts-and-correctness} in suppl. material for further detail). This approach follows recent work in evaluation frameworks \cite{vayani2024all, narnaware2025sb}, though the \textit{correctness} used here is specific to our setup.

Fig. \ref{fig:heatmap_results} shows the per-language performance comparison of different video LMMs on ViMUL-Bench. The closed-source proprietary model, GPT-4o \cite{2024GPT4o}, consistently outperforms open-source models. Among open-source models, our multilingual VidLMM, ViMUL, achieves a better tradeoff with respect to high-and low-resource languages with an overall accuracy of 51.1\%, followed by LLaVA-OneVision \cite{llava_onevision} with 49.1\%. 
Both open-source and closed-source models face challenges with several low-resource languages, such as Sinhala, Tamil, and Urdu. For example, GPT-4o's performance drops significantly from 63.2\% on English to 49.6\% on Urdu. Similarly, LLaVA-OneVision \cite{llava_onevision} drops from 65.1\% on English to 24\% on Sinhala, indicating that the model struggles with under-represented languages. In comparison, ViMUL outperforms LLaVA-OneVision by approximately 9.2\% on Urdu, and 7.5\% on Sinhala, showing its efficacy on these low-resource languages.

Fig. \ref{fig:open_vs_close} presents the performance breakdown of video LMMs on low-resource and high-resource languages. ViMUL-Bench includes three low-resource languages (Sinhala, Urdu, and Tamil), as defined in \citet{costa2022no}. The results show that the performance gap between open-source and closed-source GPT-4o increases on low-resource languages. 

\begin{figure}[!t]
    \centering
    \vspace{0.6em}
    \includegraphics[width=0.47\textwidth]{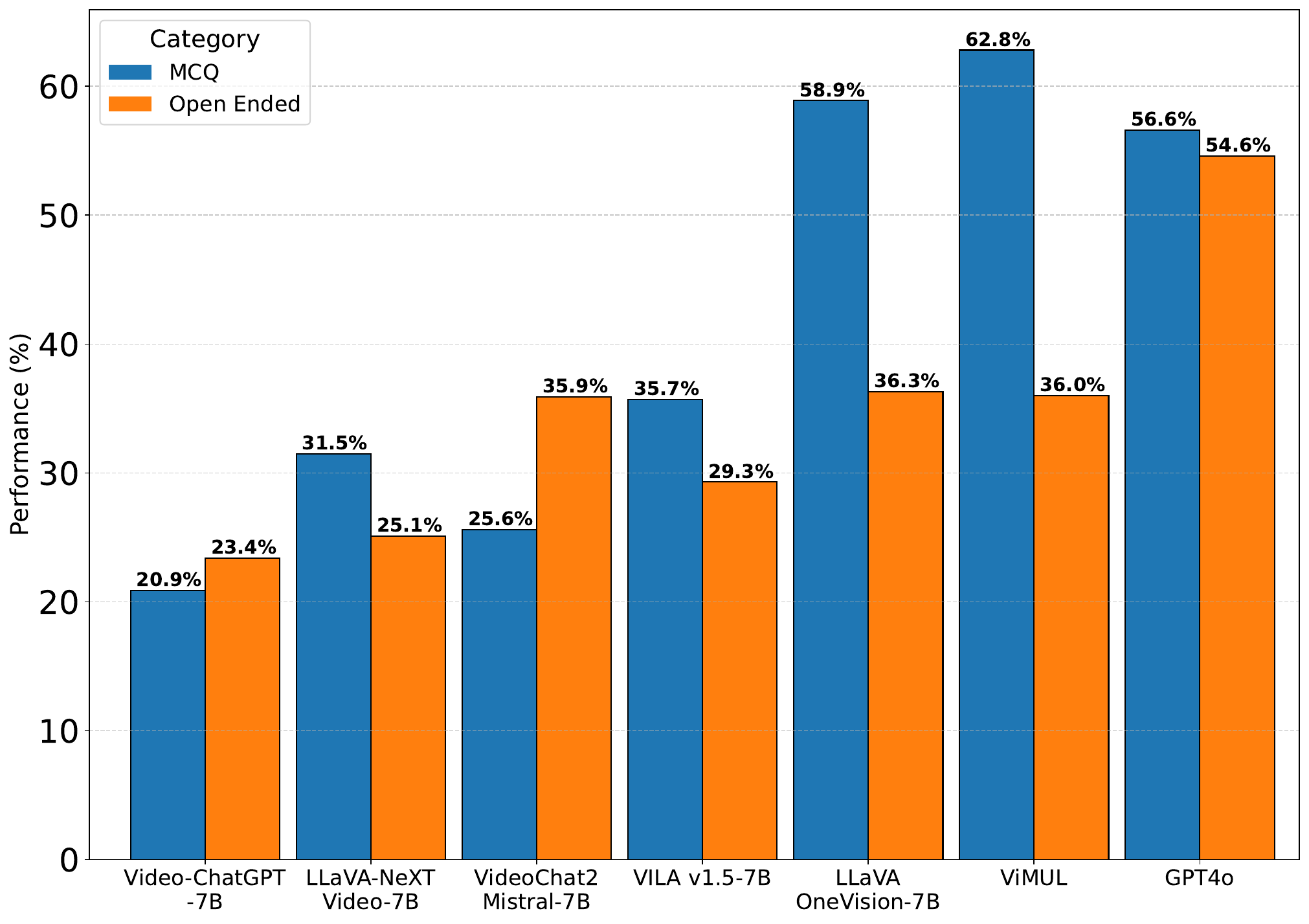}
    \vspace{-0.5em}
    \caption{\textbf{Performance of different question types in ViMUL-Bench}. Overall, the MCQs attain better performance than the open-ended QAs. ViMUL shows competitive results among open-source models.}
    \vspace{-1.5em}
    \label{fig:question_Types}
\end{figure}

\noindent\textbf{Effect of Question Type.} As mentioned earlier, ViMUL-Bench includes two types of questions: multiple-choice (MCQs) and open-ended (OE), with the latter further divided into long and short VQAs. Fig. \ref{fig:question_Types} shows the performance of video LMMs on these question types. Overall, video LMMs perform better in MCQs but struggle to generate \textit{correct} responses for OE questions. This is likely because OE questions are more complex and require enhanced understanding and multilingual reasoning across both generic and cultural domains. The only exception is VideoChat2 \cite{li2023videochat} and VideoChatGPT \cite{Maaz2023VideoChatGPT}, which perform better on OE questions than on MCQs. Among all models, the closed-source GPT-4o achieves the highest accuracy on OE questions (54.6\%), followed by ViMUL, which leads among open-source models with 36\%. Notably, ViMUL achieves the highest accuracy on MCQs (62.8\%) among all methods. We also demonstrate the consistency of Phi-4 scores with human judgement on GPT-4o outputs in Sec. \ref{phi_judge} (suppl. material).

\begin{figure}[!t]
    \centering
    \includegraphics[width=0.5\textwidth]{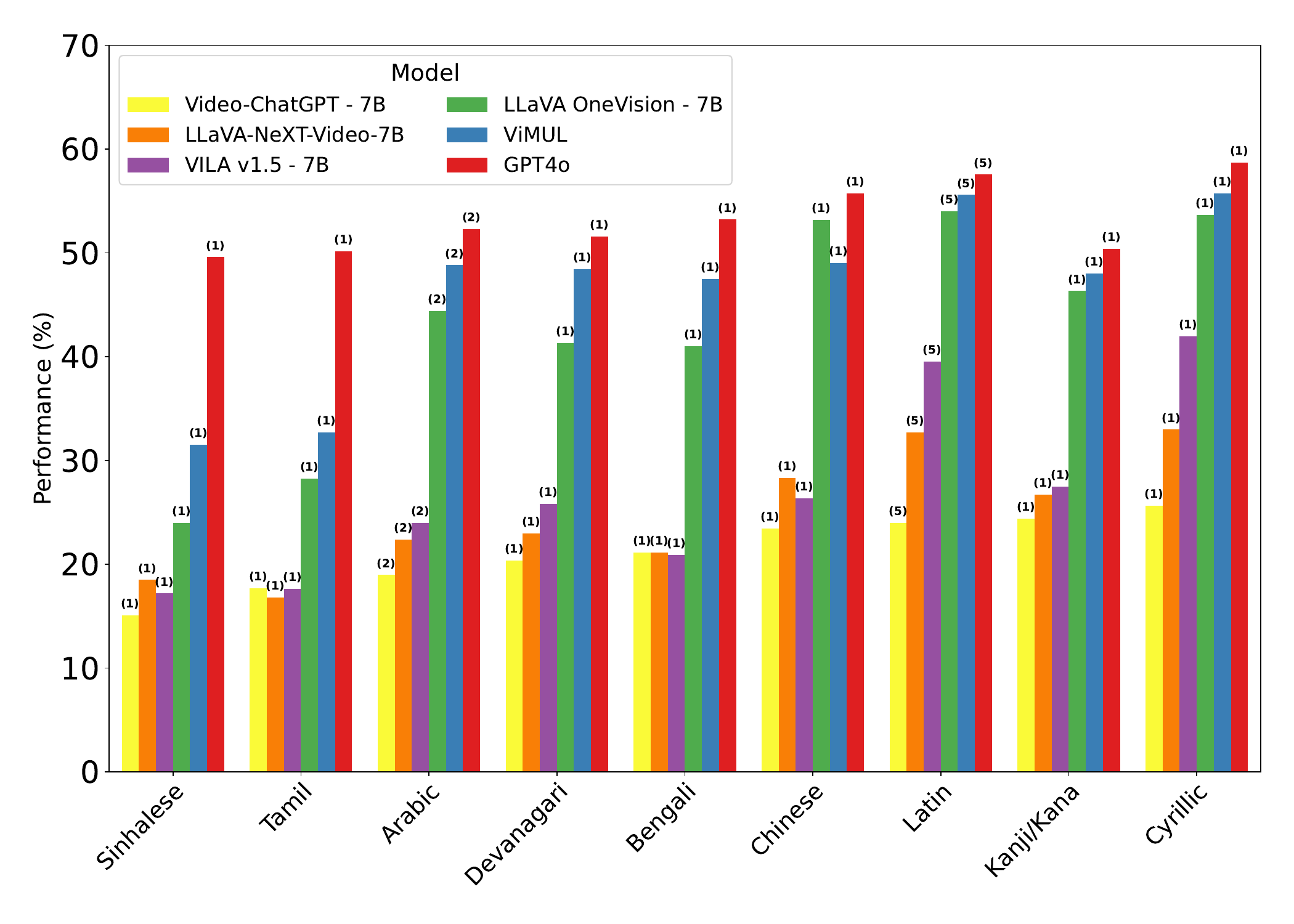}
    \vspace{-2.5em}
    \caption{\textbf{Performance on different language scripts in ViMUL-Bench}. Models fare higher on high-resource language scripts, such as Latin and Cyrillic, and Chinese, but struggle with under-represented language scripts, such as Tamil and Sinhalese.}
    \vspace{-1.5em}
    \label{fig:language_scripts}
\end{figure}

\noindent\textbf{Performance across Language Scripts.} We group the 14 languages in our ViMUL-Bench by language scripts \cite{costa2022no}, using data from Ethnologue \cite{campbell2008ethnologue} and the Glottolog database \cite{glottolog2022}. This results in nine distinct scripts. Fig. \ref{fig:language_scripts} shows the performance of video LMMs across these scripts. The closed-source GPT-4o consistently achieves the best results across all scripts. Among open-source models, ViMUL performs favorably against all existing methods across all language scripts, except Chinese, with accuracy ranging from ~32\% to ~55\%. 
Additionally, all video LMMs struggle with \textit{Sinhalese} (Sinhala) and \textit{Tamil} scripts, with ViMUL outperforming the second-best open-source model by ~8\%. A performance gradient is observed, with video LMM performing significantly better on \textit{Latin}, \textit{Chinese}, and \textit{Cyrillic} scripts compared to \textit{Sinhalese} (Sinhala) and \textit{Tamil}. 

We perform an error analysis on cultural examples from ViMUL-Bench by selecting one high-resource language \textit{(Bengali)} and one low-resource language \textit{(Sinhala)}, representing the Bengali and Sinhalese scripts, respectively. Native speakers review the open-ended subset responses within the cultural category generated by GPT-4o. Errors are categorized into seven types: lack of knowledge, lack of cultural understanding, language errors, reasoning errors, perceptual errors, translation errors \cite{vayani2024all, yue2024mmmu}, and prior-knowledge bias. We introduce prior-knowledge bias as a new error type, where the model uses prior knowledge to answer a question with information not present in the video. 

Fig. \ref{fig:pie_chart} (suppl. material) summarizes the distribution of these error types across \textit{Bengali} and \textit{Sinhalese} scripts, showing that the main errors are knowledge gaps, cultural understanding gaps, prior-knowledge bias, and lack of reasoning capabilities. Fig. \ref{fig:qual_error_types} presents error examples from language scripts. 
For instance, in \textit{Bengali}, a question about goal type is asked where GPT-4o incorrectly predicts it as a corner kick, despite the video showing a short throw-in and a header. 

\noindent\textbf{Performance Comparison across Categories.} We study the generic and cultural understanding on 15 categories  (Fig. \ref{fig:all_category_plot}). Overall, GPT-4o achieves best results of 55.8\%, but its accuracy varies significantly across different domains. For instance, it scores 75.14\% on Festivals but drops to 49.76\% on Artistic Performance. 
In contrast, categories such as \textit{Digital Content, Knowledge, Sports Competitions, and Surveillance} require a deeper understanding of visual content, likely leading to lower overall performance. 
Notably, ViMUL surpasses GPT-4o in Media \& Entertainment, achieving 57.68\%. Similarly, ViMUL outperforms GPT-4o in Notable Public Figures (56.98\% vs. 55.87\% for GPT-4o) and Sports (70.23\% vs. 68.01\% for GPT-4o). ViMUL achieves favorable performance on different categories: It achieves 43.29\% in \textit{Knowledge} and 45.78\% in \textit{Digital Content}, thereby being competitive with GPT-4o. These results show that ViMUL serves as a strong baseline for multilingual, culturally-diverse video understanding.

\noindent\textbf{Assessing the Need for a Multilingual Video Benchmark.} To motivate the design of ViMUL-Bench, we conduct three baseline ablations demonstrating the necessity of multilingual video input for fair model evaluation. \textit{(1) Blind Baseline:} We show that removing the visual input significantly degrades performance, highlighting the importance of input videos for a fair assessment. \textit{(2) Image-Only Baseline:} LMMs, when evaluated using only single frames (\textit{first/middle/last/random}), exhibit substantial performance drops compared to evaluations using the full video (32 frames), indicating that image-based LMMs are insufficient for capturing the spatio-temporal dynamics required in video benchmarks. \textit{(3) Performance on a Controlled Benchmark:} We assess the impact of our multilingual ViMUL-Instruct fine-tuning on the controlled CVRR-ES~\cite{khattak2024good} benchmark. Models show notable gains in spatio-temporal understanding, particularly in categories requiring cultural or social reasoning. Full experimental results and visualizations are provided in the suppl. material (Sec. \ref{need-for-multilingual}, Fig. \ref{fig:language_only}, Fig. \ref{fig:finetuning_before_after}, Tab.~\ref{tab:frame_experiment}).

\noindent\textbf{Impact of Location-aware Information in Prompts:} 
Tab. \ref{tab:country_info_performance} (suppl. material) presents the comparison when provided with additional country-specific information. Results are consistent with observations in \citet{vayani2024all}, with improvement due to better utilization of geographic context.

\noindent\textbf{Impact of Video Duration.} We further group the videos in ViMUL-Bench into three broad categories based on their duration: short, medium, and long, and present our results in Fig. \ref{fig:video_durations} (suppl. material). Overall, GPT-4o outperforms other models on short and medium videos. However, ViMUL surpasses GPT-4o and other methods on long videos in the multilingual setting. Further details are provided in Sec. \ref{video-durations} (suppl. material).

%% file: sec/7_conclusion.tex
\section{Conclusion}

We introduce ViMUL-Bench, the first multilingual benchmark explicitly designed to evaluate video LMMs across diverse linguistic and cultural scenarios. It comprises over 8k humanly verified QA pairs across 14 languages, including both high-resource and several low-resource languages and spanning 15 diverse categories. Further, we present a large-scale multilingual video instruction tuning dataset comprising 1.2 million samples, which we use to develop a simple multilingual video LMM demonstrating competitive cross-linguistic and cultural comprehension.

\section*{Acknowledgement}
The computations were enabled through resources provided by NAISS at Alvis, partially funded by the Swedish Research Council under grant agreement no. 2022-06725, LUMI, hosted by CSC (Finland) and the LUMI consortium, and the Berzelius resource provided by the Knut and Alice Wallenberg Foundation at the NSC. This work was partially supported by the Swedish Research Council (2022-04266), from KAW (DarkTree project; 2024.0076) and VR starting grant (2016-05543).

%% file: sec/8_limitations.tex
\section{Limitations}
ViMUL, to the best of our knowledge, is the first multilingual Video LMM benchmark that exhibits cultural and linguistic inclusivity across 14 languages and 15 diverse domains. However, despite being the first benchmark for evaluating Video LMMs in a multilingual setting, it has certain limitations.  ViMUL-Bench contains more general VQA samples in its sub-categories than in the cultural section, largely due to the high cost of curating culturally grounded data, which demands extensive human effort. It also has more short and medium-length videos than long ones, as high-quality long videos are harder to source for low-resource languages such as Sinhala, Tamil, and Urdu, and require extra verification for reliable QA pairs. While we applied cycle consistency checks during multilingual instruction fine-tuning, further human validation could improve quality. Finally, the ViMUL model, though a simple baseline, highlights the trade-off between high- and low-resource languages in video understanding. As future research directions, our work can be extended to include additional languages, particularly those with limited digital representation. Furthermore, developing culturally-specific, large-scale training datasets tailored explicitly to underrepresented communities would further enhance the inclusivity and effectiveness of multilingual video LMMs. We also expect culture-specific video-language preference data collection can help improve LMMs' performance further via RL. 

%% file: sec/6_ethical_consideration.tex
\section{Ethical Consideration}
Our work reports a standardized multilingual video LMM evaluation benchmark. We hope both ViMUL and ViMUL-Bench will contribute to more consistent evaluation across diverse domains, particularly for underrepresented languages in VidLMM research. Since the cultural videos in ViMUL-Bench are sourced from Internet, some domains may be under-represented, leading to potential biases. To ensure high translation quality, fluent native speakers thoroughly reviewed and verified our initial GPT-4o-generated translations for consistency and accuracy. The verification involves 16 volunteers from diverse linguistic backgrounds, requiring having familiarity with the cultural context of the specific country-language pair they worked on. Additional annotator demographic details are presented in Sec. \ref{volunteerDemographics_z} suppl. material.

%% file: sec/X_suppl.tex
\clearpage

\section{Prompts used in Evaluation}
\label{promts-and-correctness}
\paragraph{Multiple-Choice Evaluation Prompt.}

All LMMs are prompted on the below multiple-choice evaluation prompt to select the most accurate answer from the available options based on both the video sample and the subtitles. The model is strictly instructed to select only one correct option (A, B, C, or D), ensuring a clear and direct response format. Then, the model's generated output is compared directly with the ground-truth to evaluate accuracy. The prompt for MCQ evaluation is shown below:

\begin{tcolorbox}[colback=blue!5!white, colframe=blue!75!black, title=LMM Inference Prompt.]

\textbf{Prompt:} \\
Select the best answer to the following multiple-choice question based on the video and the subtitles. Respond with only the letter (A, B, C, or D) of the correct option.
\end{tcolorbox}

\paragraph{Open-Ended Evaluation Prompt.} Below, we list the evaluation guidelines for the open-ended evaluation.

\begin{tcolorbox}[colback=blue!5!white, colframe=blue!75!black, title=Evaluation Guidelines]

\textbf{System Prompt:} \\
You are an intelligent chatbot designed for evaluating the correctness of AI assistant predictions for question-answer pairs. 
Your task is to compare the predicted answer with the ground-truth answer and determine if the predicted answer is correct or not. Here's how you can accomplish the task:

\vspace{5pt}
\textbf{Instructions:}
\begin{itemize}
    \item \textbf{Correctness Focus}: Compare the predicted answer with the ground-truth answer to determine accuracy.
    \item \textbf{Detail Consideration}: Predictions with fewer details are still correct unless such details are explicitly required in the question.
    \item \textbf{Scoring}: Assign an integer score between 0 (fully incorrect) and 5 (fully correct), with intermediate values reflecting partial correctness.
\end{itemize}
\end{tcolorbox}

This GPT prompt is designed for evaluating AI assistant predictions on video-based question-answer pairs. It instructs the model to compare predicted responses with ground-truth answers, assessing correctness while allowing for minor variations unless explicitly required. The evaluation follows a structured scoring system from 0 (fully incorrect) to 5 (fully correct), with intermediate scores reflecting partial accuracy. The output is formatted as a Python dictionary containing the prediction status ("correct" or "incorrect"), a numerical score, and a justification, ensuring consistency in automated assessment for multilingual and multimodal AI systems.

\begin{tcolorbox}[colback=blue!5!white, colframe=blue!75!black, title=Evaluation Request.]

Please evaluate the following video-based question-answer pair:

\textbf{Question}: \{question\} \\
\textbf{Ground Truth Correct Answer}: \{answer\} \\
\textbf{Predicted Answer}: \{pred\} 

Provide your evaluation as a correct/incorrect prediction along with the score, which is an integer value between 0 (fully wrong) and 5 (fully correct). The middle score represents the percentage of correctness.

\vspace{5pt}
\textbf{Response Format:} \\
Your response should be generated as a Python dictionary string with the following keys:
\begin{itemize}
    \item \textbf{`pred'}: A string, either ``correct" or ``incorrect".
    \item \textbf{`score'}: An integer between 0 and 5.
    \item \textbf{`reason'}: A justification for the decision.
\end{itemize}

Only provide the Python dictionary string. \\ \\
Example format:
\begin{verbatim}
{"pred": "correct", "score": 4, 
"reason": "The predicted answer 
captures most of the ground-truth 
meaning but lacks minor details."}
\end{verbatim}

\end{tcolorbox}

\begin{figure*}[t]
  \centering
    \includegraphics[width=1.0\linewidth]{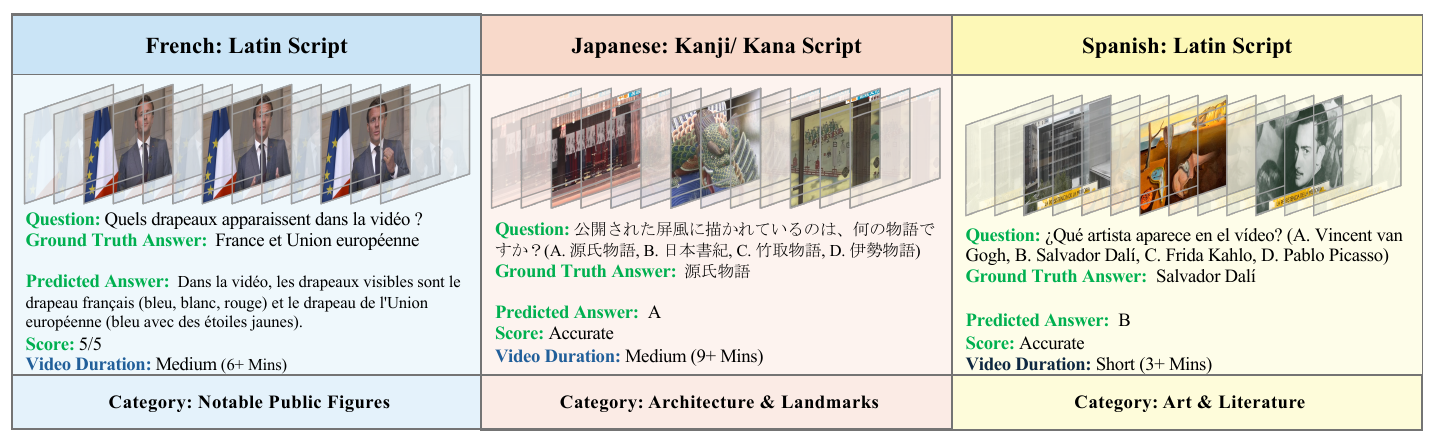}\vspace{-0.8em}
    \caption{\small We present qualitative examples of success cases of GPT-4o’s across different language scripts and categories.}
    \label{fig:qual_success_cases}

    \vspace{-1em}
\end{figure*}

\section{Dataset Statistics}
\begin{figure}[!h]
    \centering
    \includegraphics[width=0.3\textwidth]{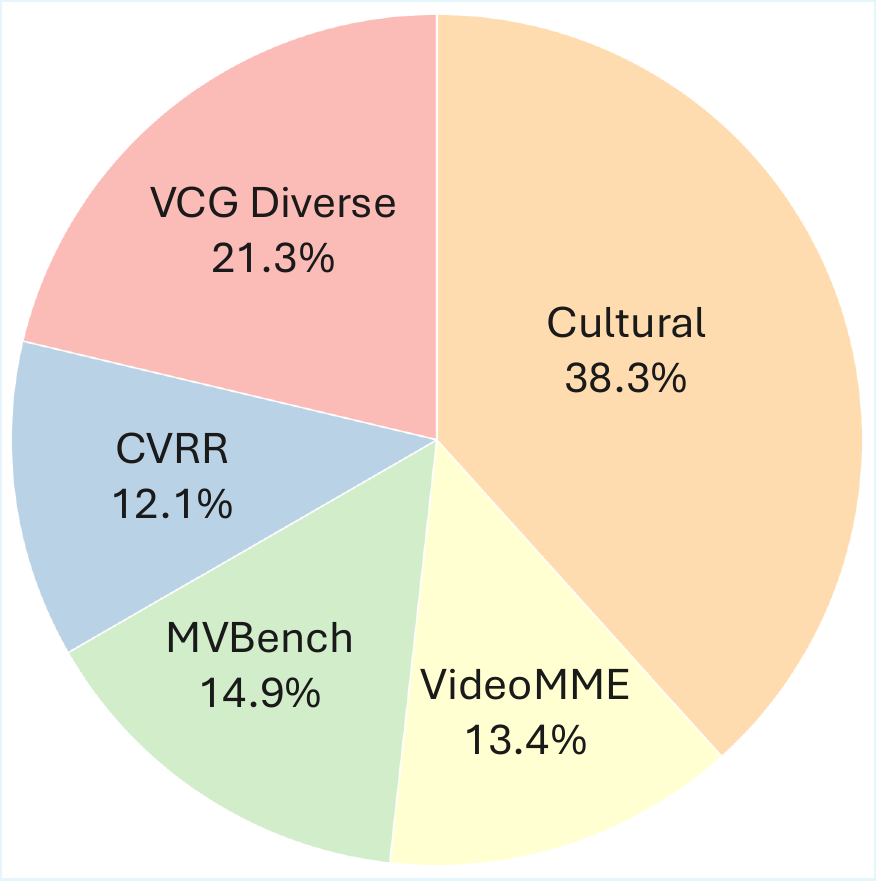}
    \caption{We present the cultural versus generic category distribution. We source generic categories from existing video benchmarks. However, the cultural part is carefully curated from scratch by native speakers.}
    \label{fig:cultural_vs_generic_distribution}
\end{figure}

\paragraph{Cultural vs Generic Distribution.}
ViMUL-Bench includes a total of 15 diverse categories, comprising both generic and cultural categories. Figure \ref{fig:cultural_vs_generic_distribution} illustrates the distribution of these categories across different sources. The generic samples are carefully selected from VCG-Diverse, CVRR-ES, MVBench, and VideoMME datasets. For the cultural category, we curate the content from scratch with input from native speakers for all languages except English. Additionally, we use GPT-4o to translate video question-answer pairs from all 13 languages into English.

\paragraph{Video Duration Distribution.}

\begin{figure}[!h]
    \centering
    \includegraphics[width=0.3\textwidth]{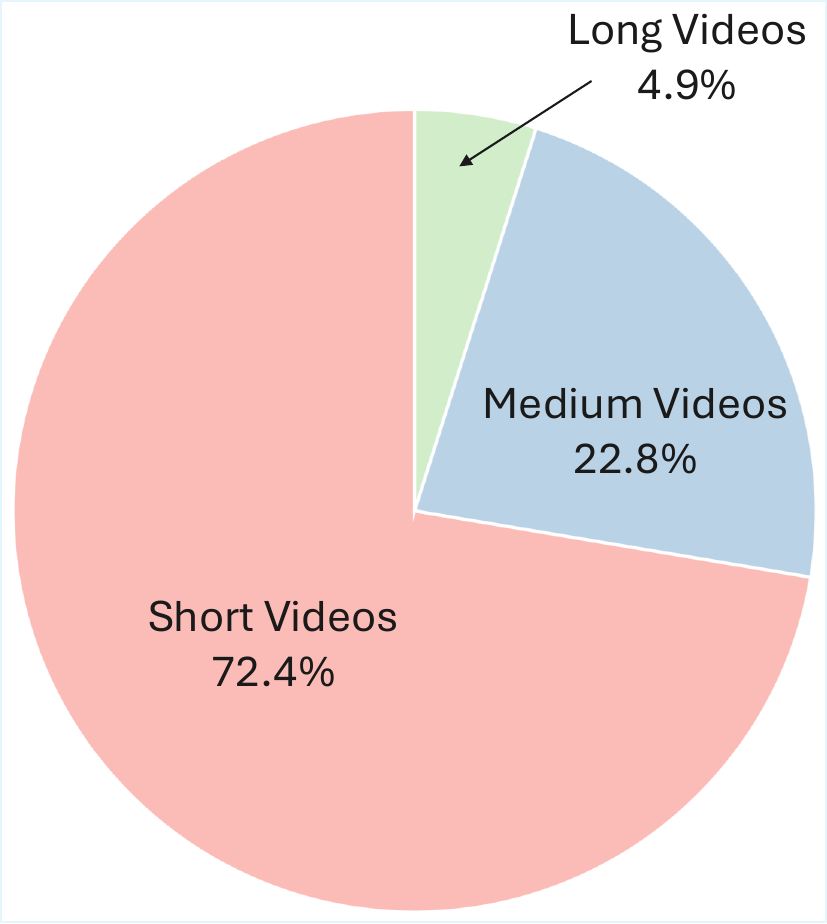}
    \caption{The figure illustrates the video duration distribution in ViMUL-Bench. Overall, we have over 73\% short-duration videos in the dataset.}
    \label{fig:video_duration_distribution}
\end{figure}

ViMUL-Bench categorizes videos into three durations: short (0-4 minutes), medium (4-15 minutes), and long (15+ minutes). Each video is manually assigned a duration label. Figure \ref{fig:video_duration_distribution} illustrates the distribution of video durations across the dataset. The plot shows that over 73.7\% of the videos are short, making it the largest category, followed by 20.5\% medium-duration videos and 5.8\% long-duration videos. Although we aimed to achieve a balanced distribution of video durations, curating long-duration videos proved to be more resource-intensive for verification, and they are relatively scarce.

\begin{figure}[!h]
    \centering
    \includegraphics[width=0.4\textwidth]{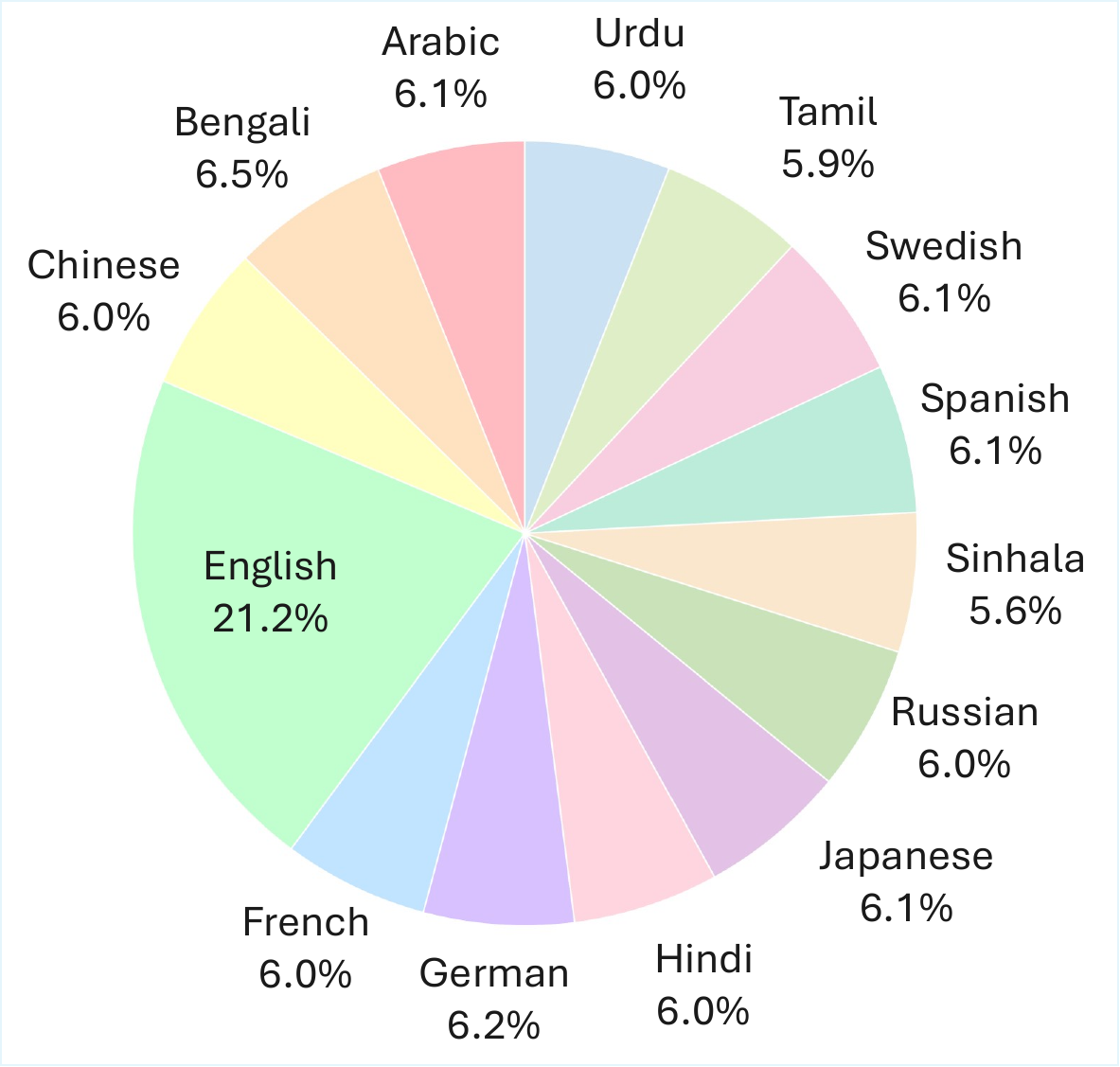}
    \caption{The figure illustrates the per-language distribution in ViMUL-Bench. The dataset contains nearly equal proportions of both low-resource and high-resource languages, except English. English comprises translations of video question-answer pairs from all 13 other languages in the dataset.}
    \label{fig:language_distribution}
\end{figure}

\paragraph{Language Distribution.}
Figure \ref{fig:language_distribution} illustrates the language distribution across ViMUL-Bench. The dataset contains a nearly equal proportion of videos for each of the 14 languages, with the exception of English, which is more heavily represented. For the cultural data, we translate video question-answer pairs from all 13 non-English languages into English to ensure consistency and facilitate cross-lingual comparisons. This approach enables us to maintain linguistic diversity while ensuring accessibility and usability of the dataset in English.

\begin{figure}[!h]
    \centering
    \includegraphics[width=0.4\textwidth]{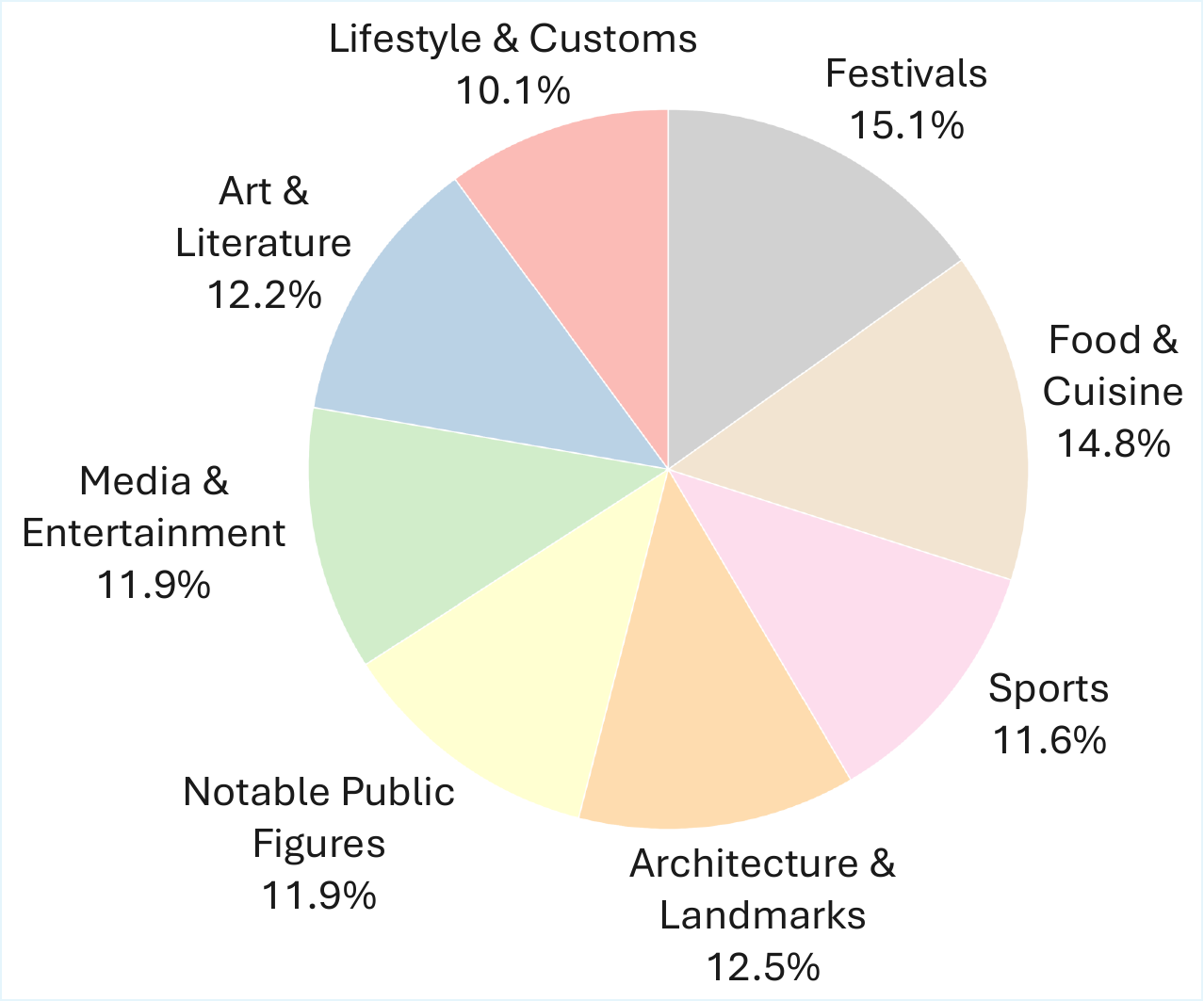}
    \caption{The figure illustrates the Distribution of the eight cultural categories in ViMUL-Bench, where we ensure consistent samples across all the cultural categories.}
    \label{fig:category_distribution_cultural}
\end{figure}

\begin{figure}[!h]
    \centering
    \includegraphics[width=0.4\textwidth]{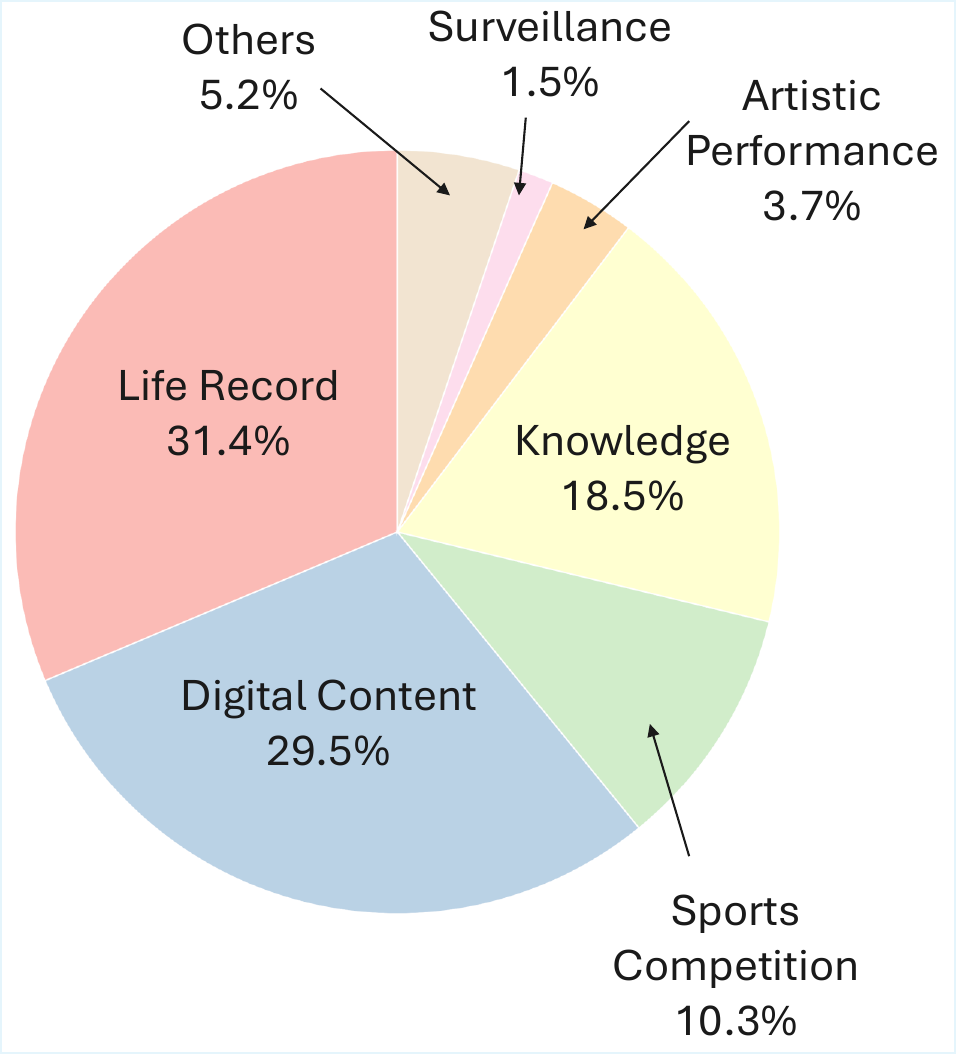}
    \caption{The figure illustrates the Distribution of the seven generic categories in ViMUL-Bench, with the largest proportions in Life Record, Digital Content, and Knowledge from generic categories.}
    \label{fig:category_distribution}
\end{figure}

\paragraph{ViMUL Category Distribution.} Our benchmark consists of 15 diverse categories, including seven generic categories and eight cultural categories. Figure \ref{fig:category_distribution} presents the distribution of the seven generic categories. Among these, the Life Record category accounts for approximately 23\% of the samples, followed by Digital Content (17.5\%) and Knowledge (11.1\%). Surveillance and Artistic Performance categories represent smaller proportions of the dataset. In total, over 33\% of the dataset is composed of cultural categories, which are further divided into eight subcategories. Fig. \ref{fig:category_distribution_cultural} shows the cultural category distribution, where all the categories are almost balanced, ranging from 15.1\% of QA pairs in Festival to 10.1\% in Lifestyle and Customs. 

\section{Impact of Location-aware Information in Prompts}
Table \ref{tab:country_info_performance} presents the performance improvements when geographical information, specifically country details, is included in the prompts. The closed-source model demonstrates a greater ability to leverage this additional information compared to the open-source model.

\begin{table}
    \centering
    \resizebox{\columnwidth}{!}{
    \begin{tabular}{lcc}
        \toprule
        \textbf{Models} & \textbf{With Country Info.} & \textbf{Without Country Info.} \\
        \midrule
        \rowcolor{LGray} GPT-4o & \textcolor{darkgreen}{63.3\%} & 60.8\% \\
        \arrayrulecolor{gray} \midrule
        \rowcolor{LGray} ViMUL & \textcolor{darkgreen}{53.1\%} & 52.61\% \\
        \arrayrulecolor{black} \bottomrule
    \end{tabular}
    }
    \caption{\textbf{Performance with and without additional country location information}. Results improve when integrating additional geographic information as input to VidLMMs.
    }
    \label{tab:country_info_performance}
\end{table}

\section{Volunteer Demographics}
\label{volunteerDemographics_z}
\begin{figure}[!h]
    \centering
    \includegraphics[width=1.1\linewidth]{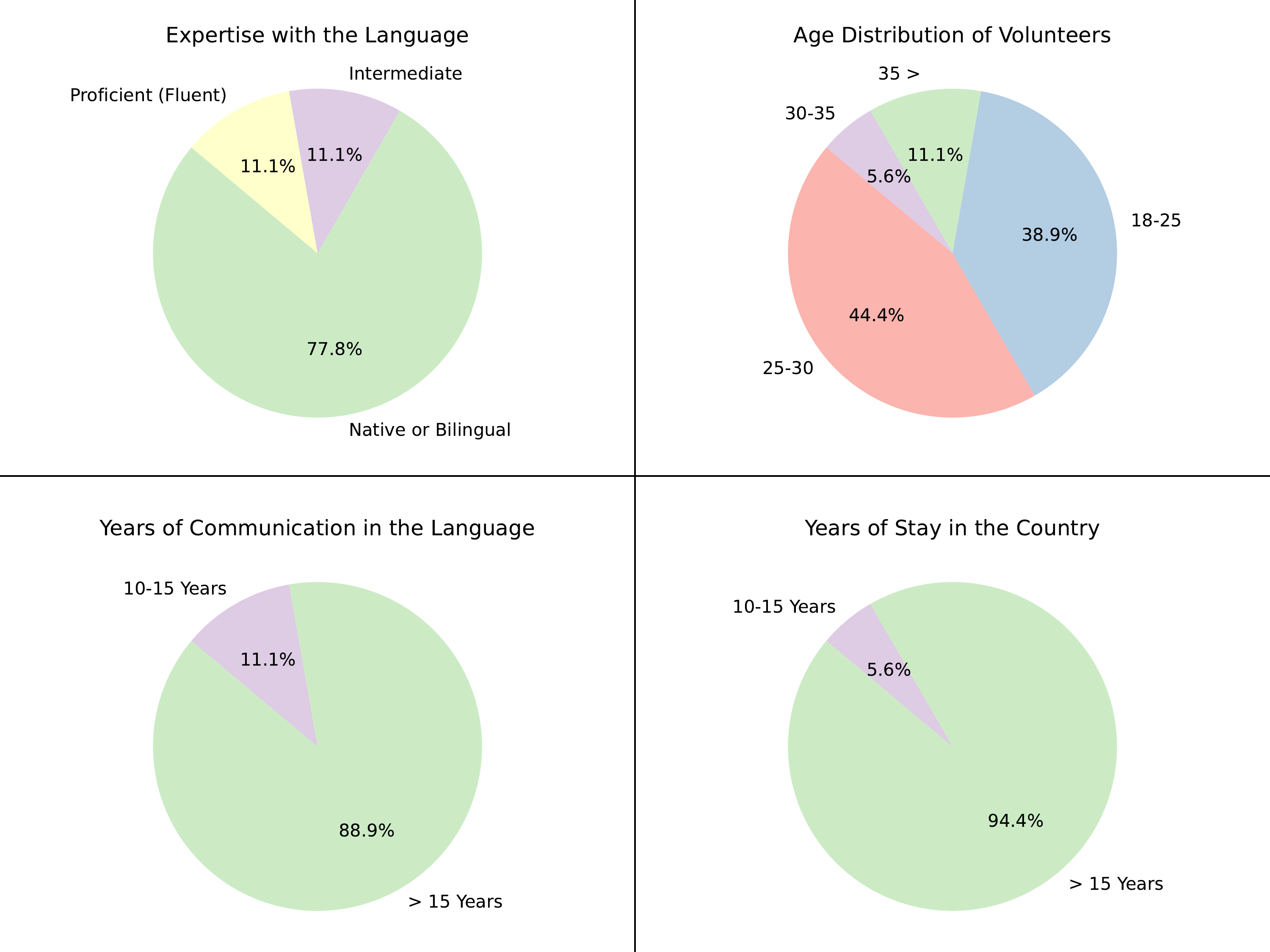}
    \caption{The top left figure shows the percentage of our volunteers with respect to linguistic skill level. The top right shows their age distribution. Then in bottom left we have their active years of communication, followed by the duration of their stay in their respective countries.}

    \label{fig:volunteers}
\end{figure}

We have a total of 16 volunteers from various backgrounds who assisted us in curating and verifying our ViMUL Bench. Among these, two volunteers are proficient in two different languages each, and they contributed separately for both languages. To accurately reflect their contributions and language proficiency levels, we consider a total of 18 language instances rather than just 16 individuals when calculating statistics. 

Around 77.8\% of our verifiers are native or bilingual, followed by 11.1\% of them proficient and the rest are intermediate in their respective languages. Around 94.4\% stayed for more than 15 years in their country, where they learned their first language. We can observe in Fig. \ref{fig:volunteers} our contributors range from the age bracket of 18 - 25, 25 - 30 and onwards, which makes the age distribution more diverse, and includes prior experience. In terms of geographical distribution, they come from the following countries as follows Bangladesh, China, Germany, India, Japan, Lebanon, Morocco, Pakistan, Sri Lanka, Sweden, Ukraine and USA. This diversity in geo-location helps us to cater to cultural nuances of that corresponding region, hence allowing us to get translations and verifications which are authentic to that place.

\section{Cycle Consistency for Train Set}

\begin{figure}[!h]
    \centering
    \includegraphics[width=1\linewidth]{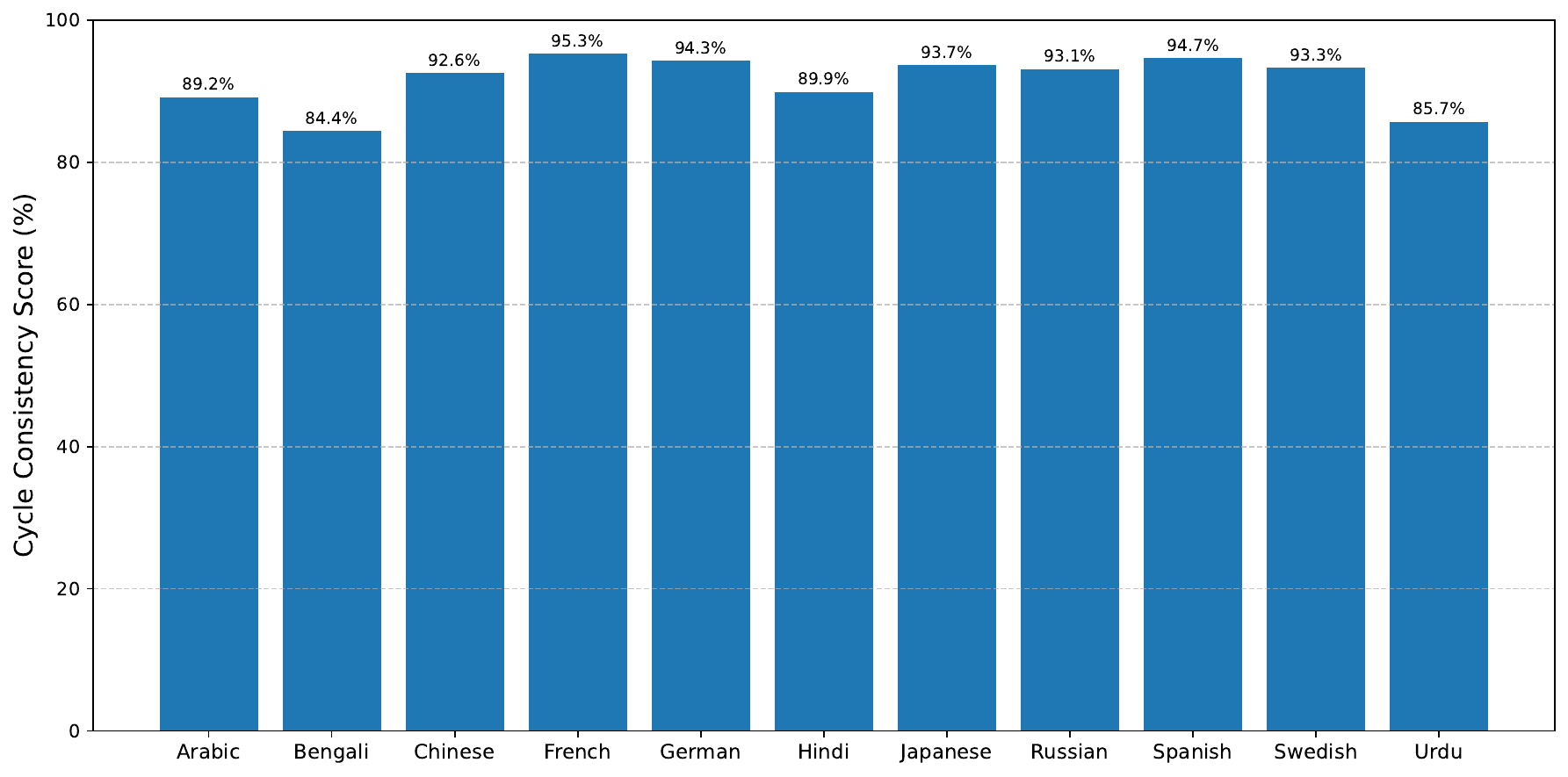}
    \caption{\textbf{Performance of Cycle consistency upon 11 languages}. All the 11 languages display an accuracy well above 80\%, with the average being 91.47\% across the languages.}

    \label{fig:cycle_consistency}
\end{figure}

We perform a cycle consistency check to evaluate the quality of machine translations across multiple languages. Starting with a pre-existing English-only training dataset, we translate the samples into 13 additional languages using GPT-4o-mini. To assess the accuracy and fidelity of these translations, we then back-translate the non-English samples into English using Qwen 3. The back-translations are evaluated based on the following criteria: consistency with the original text, grammatical correctness, and meaningfulness. These evaluations are conducted using GPT-4o, allowing us to systematically score and compare machine translation in terms of both linguistic and semantic properties between languages. We can observe in Fig. \ref{fig:cycle_consistency} that the highest performance can be seen in French with 95. 3\%, followed by Spanish with 94.7\%. At the lower end, we can find Bengali with 84.4\%, followed by Urdu with 85.7\%.   

\section{Performance of Phi-4 as a judge}

\begin{figure}[!h]
    \centering
    \includegraphics[width=1\linewidth]{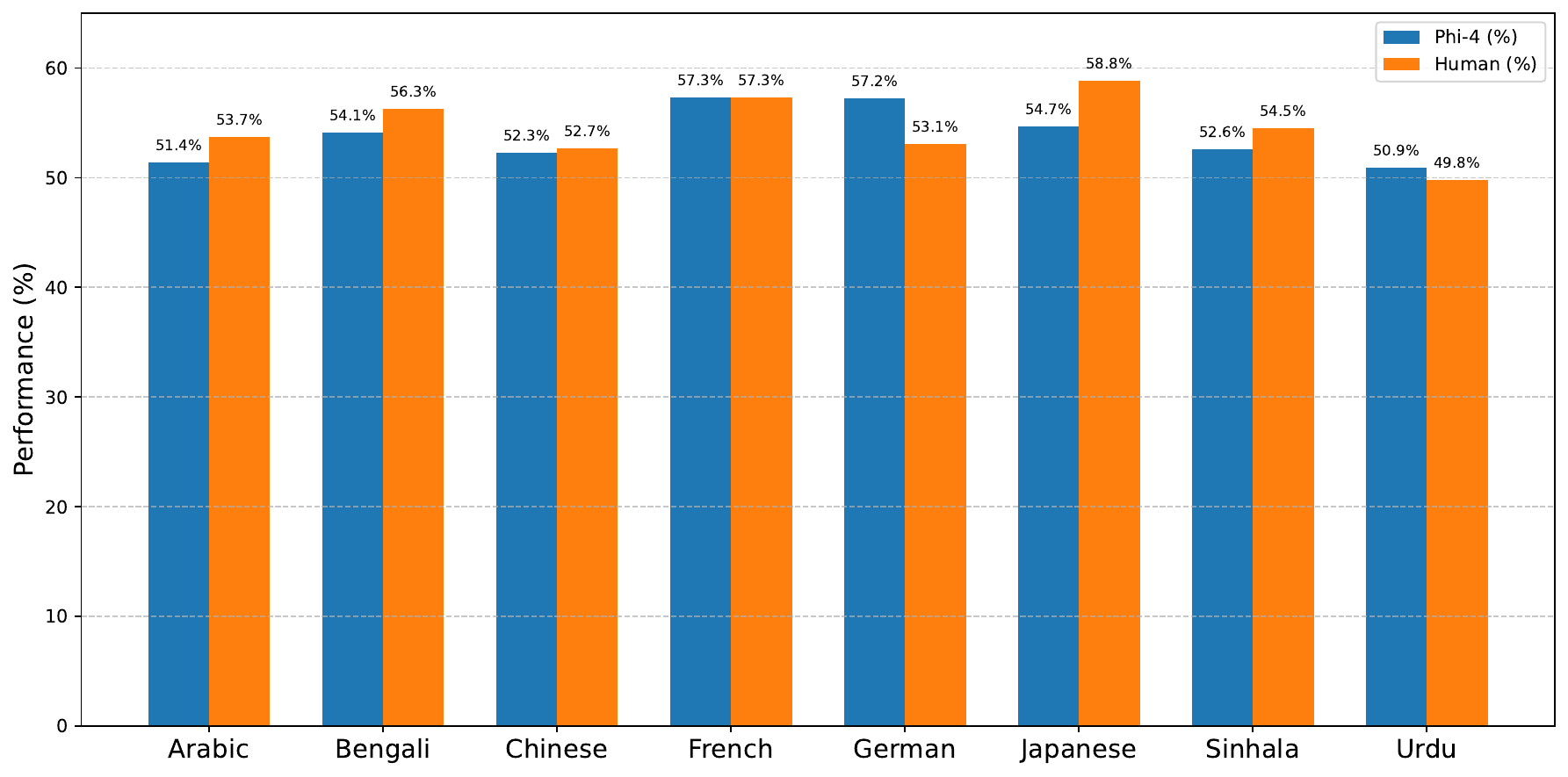}
    \caption{\textbf{Performance of Phi-4 scores compared to human score}. The figure illustrates that Phi-4 scores are similar to human scores where they differ most in Japanese by 4.1\% and the least in Chinese by 0.4\%.}

    \label{fig:human_gpt_scores}
\end{figure}

\label{phi_judge}
To validate the reliability of Phi-4 as a judge, we conducted human verification on 100 randomly sampled open-ended VQAs scores (both cultural and generic categories) from our ViMUL-Bench. Fig. \ref{fig:human_gpt_scores} (suppl.) shows the performance consistency on GPT-4o responses, where all languages, both low resource and high resource languages, exhibit strong agreement with Phi-4's scores when compared to manual native speaker scores. For instance, for Urdu language, the average accuracy of Phi-4 score is 50.9\%, whereas for human score the average is 49.8\%. Hence we can deduce that since the Phi-4 scores are similar to human scores, thus Phi-4 scores are reliable.

\section{Performance of model based upon question only}

\begin{figure}[!h]
    \centering
    \includegraphics[width=1\linewidth]{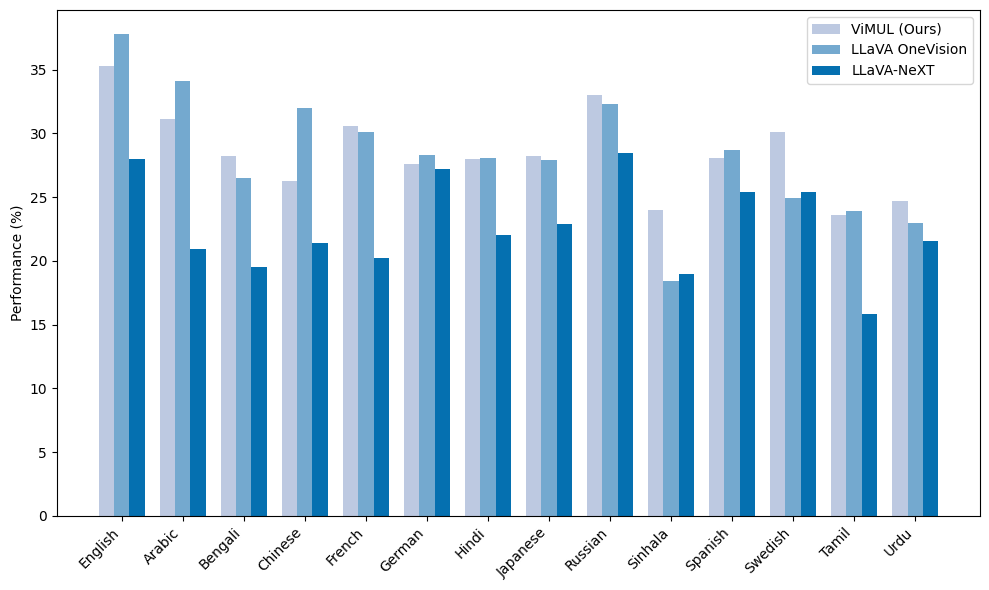}
    \caption{\textbf{Performance of models for question only as input}. This Figure shows the performance of models for all 14 languages when only question is given as input [No frames].}

    \label{fig:language_only}
\end{figure}

Fig. \ref{fig:language_only} compares the performance of ViMUL, LLaVA OneVision, and LLaVA-NeXT across 14 languages when only the question is provided as input (no frames). The results highlight that ViMUL achieves a slightly higher average performance of 28.5\%, outperforming LLaVA OneVision, which has an average of 28.3\%, and LLaVA-NeXT, which averages 22.7\%. ViMUL demonstrates a better overall tradeoff between high-resource and low-resource languages. Specifically, in languages such as Sinhala and Urdu, which are considered low-resource, ViMUL maintains relatively stronger performance compared to LLaVA OneVision and LLaVA-NeXT. This shows that ViMUL is more robust across both high-resource languages (like Russian, French and Swedish) and low-resource languages, making it a more versatile model for multilingual tasks.

\section{Performance of ViMUL-LLM before and after finetuning}

\begin{figure}[!h]
    \centering
    \includegraphics[width=1\linewidth]{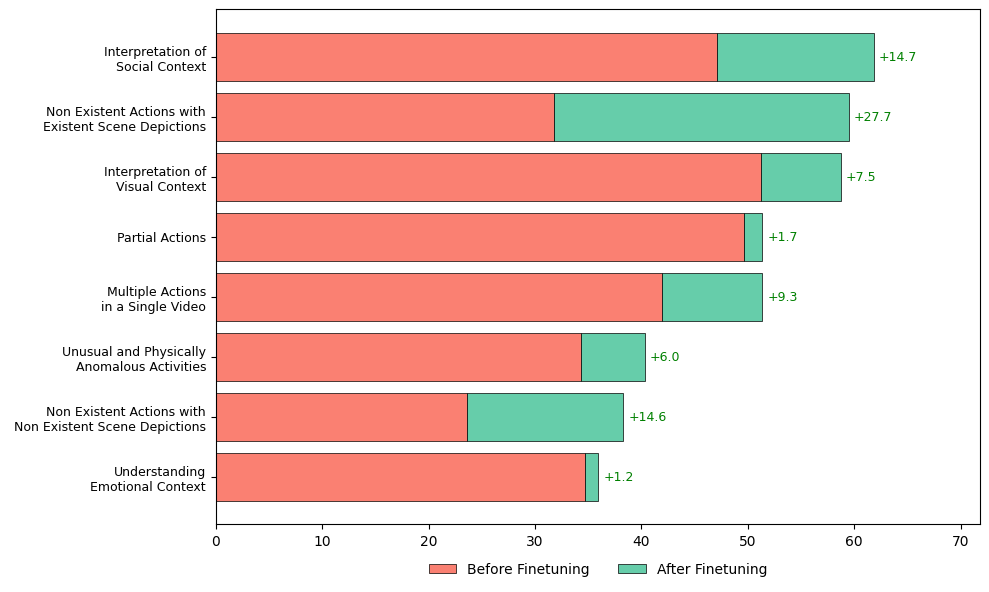}
    \caption{\textbf{Performance before Vs after finetuning}. This Figure portrays the performance enhancement of ViMUL model after finetuning for various tasks, such as spatial and temporal.}

    \label{fig:finetuning_before_after}
\end{figure}

Fig. \ref{fig:finetuning_before_after} demonstrates the performance of the ViMUL model before and after fine-tuning across eight distinct tasks. The tasks include a variety of challenges such as interpreting social and visual contexts, handling multiple actions in a single video, and recognizing emotionally charged or anomalous activities. The figure shows the improvement in performance after fine-tuning, with the score for each task before and after fine-tuning represented by horizontal bars. The difference in scores, indicated by the green bars, highlights the model's enhancement across different tasks, particularly in areas like \textit{Interpretation of Social Context} and recognizing \textit{Non Existent Actions with Existent Scene Depictions}, which increases by 14.7\% and 27.7\%, respectively. This visual comparison underscores the effectiveness of the fine-tuning process in boosting the overall accuracy of the model and the specific performance of the task.

\section{Performance based upon frames}

\begin{table}[h]
\centering
\small
\setlength{\tabcolsep}{3.5pt}  
\begin{tabular}{p{2.3cm}ccc}
\toprule
\textbf{Setting} & \textbf{LLaVA-OneVision} & \textbf{ViMUL (Ours)} \\
\midrule
First Frame   & 34.1 & 35.3 \\
Mid Frame     & 36.7 & 39.0 \\
Last Frame    & 34.6 & 35.8 \\
Random Frame  & 37.7 & 38.4 \\
\midrule
ViMUL-Bench   & 46.2 & 49.2 \\
\bottomrule
\end{tabular}
\caption{Comparison of image and video backbones. Accuracy of LLaVA-NeXT, LLaVA OneVision, and ViMUL across different frame inputs.}
\label{tab:frame_experiment}
\end{table}

Tab. \ref{tab:frame_experiment} compares the performance of LLaVA-OneVision and ViMUL model across different input types, including first, mid, last, and random frames, as well as when 32 frames are taken as input, referred to as "ViMUL-Bench." The results demonstrate that ViMUL consistently outperforms LLaVA-OneVision across all frame-based settings. Notably, ViMUL achieves the highest accuracy of 49.2\% in the ViMUL-Bench setting, where 32 frames are used as input. This significant performance boost underscores the importance of videos over individual images. Videos provide crucial temporal context that images alone cannot capture, allowing models to leverage the dynamics and sequences of actions, resulting in better overall accuracy. The improvement observed with multiple frames emphasizes that incorporating temporal information in video inputs is key to enhancing model performance.

\section{ViMUL: Multilingual Video LMM}
\label{mutlilingual_videolmm}

\noindent
\textbf{Video Sampling:} 
Given an input video $ \mathbf{V} \in \mathbb{R}^{T \times H \times W \times C} $, where $T$ is the total number of frames, and $H$ and $W$ denote the height and width of each frame respectively, we define $N$ as the maximum number of frames that can be processed.  We first sample the video at 1 FPS, resulting in $t$ frames. If $t \leq N$, we keep all sampled frames. However, if $t > N$, we further uniformly select $N$ frames from the $t$ frames. This ensures that the final number of frames does not exceed $N$. Thus, the video representation after sampling is  
$
\mathbf{V'} \in \mathbb{R}^{n \times H \times W \times C}
$, where $n = \min(N, t)$ and $\mathbf{V'}$ represents the sampled video with at most $N$ frames.

\noindent
\textbf{Video Encoding:}
We use SigLIP~\cite{siglip} as the vision encoder, followed by a vision-to-language projector to transform vision tokens into the input embedding space of the language model. Given the video $\mathbf{V'} \in \mathbb{R}^{n \times H \times W \times C}$, we first resize the input to the resolution that SigLIP is trained on, followed by encoding each frame of the video using SigLIP. The output from the second-to-last layer of SigLIP is flattened and passed through a two-layer MLP, which projects the vision features into the language model's embedding space. The projected features are reshaped into a grid format and pooled using a $2 \times 2$ kernel to reduce the number of features by a factor of four. Empirically, we found that this helps accommodate more video frames while maintaining performance. Finally, the projected features are flattened, resulting in video embeddings $\mathbf{E}^{vid} \in \mathbb{R}^{n \times L_v \times D_t}$, where $L_v$ represents the total visual features per video frame, and $D_t$ is the embedding size of the language model.

\noindent
\textbf{Language Model}: 
We obtain the final representation by concatenating the video embeddings $\mathbf{E}^{vid}$ with the text embeddings $ \mathbf{E}^{text} \in \mathbb{R}^{L \times D_t} $ of the user query,
\begin{equation}
    \mathbf{E} = [\mathbf{E}^{vid},  \mathbf{E}^{text}].
\end{equation}
This ensures the language model receives spatio-temporal video features followed by the user query to generate an accurate response. We use Qwen-2.0~\cite{qwen2blog} as the language model and fully fine-tune it in an auto-regressive manner with a next-token prediction loss (see Fig.~\ref{fig:mint_lmm_block_dia}).

\textbf{Dataset Distribution}: 
The LLaVA-Video-178K dataset includes 178,510 caption entries, 960,792 open-ended QA pairs, and 196,198 multiple-choice samples. We employ LLaMA-3.1-70B-Instruct~\cite{llama3} model to identify the complex QA pairs in both open-ended and multiple-choice categories. Specifically, we prompt the LLM to identify samples that are difficult and require chain-of-thought reasoning, resulting in 39,422 QA pairs. Further, we include the training sets of NeXT-QA~\cite{xiao2021next}, PerceptionTest~\cite{patraucean2023perception}, and Clevrer~\cite{yi2019clevrer}, contributing an additional 29,846 samples. This results in a total of 95k samples in our English training dataset.

We translate these English QA pairs into 13 other languages, including Arabic, Bengali, Chinese, French, German, Hindi, Japanese, Russian, Sinhala, Spanish, Swedish, Tamil, and Urdu, using GPT-4o-mini~\cite{2024GPT4o}. We observe that GPT-4o occasionally makes obvious mistakes in translation, such as failing to respond in the target language. To address these issues, we employ LLaMA-3.1-8B~\cite{llama3} model to post-process the translations. We input the translated QA pairs to the LLM and ask it to predict the language in which the text is written. If the LLM predicts that the text is not in the intended language, we simply discard these samples.

Finally, our pipeline results in a total of 1,238,102 samples across all languages. Original English dataset contains 95,071 samples, while translated datasets contain; Arabic: 88,154, Bengali: 88,087, Chinese: 88,095, French: 86,990, German: 88,020, Hindi: 88,004, Japanese: 88,041, Russian: 88,054, Sinhala: 88,023, Spanish: 87,976, Swedish: 87,974, Tamil: 87946, Urdu: 87667 samples, respectively.

\begin{figure}[!h]
    \centering
    \includegraphics[scale=0.5, width=0.47\textwidth]{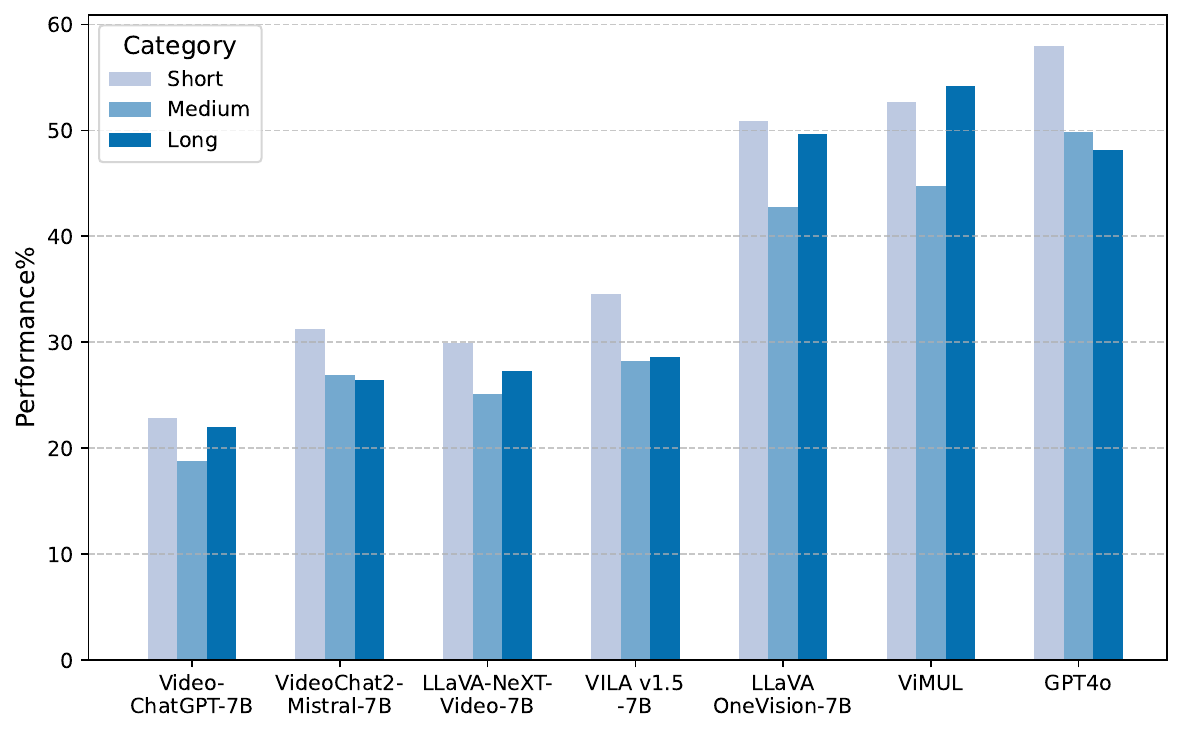}
    \vspace{-1em}
    \caption{\textbf{Performance comparison across three video durations in ViMUL-Bench}. Most methods perform better on short video questions, followed by long and then medium-length videos. GPT-4o achieves the highest accuracy on short VQA, while ViMUL outperforms all LMMs on long VQAs.}
    \label{fig:video_durations}
\end{figure}

\section{Impact of Video Duration}
\label{video-durations}

As mentioned earlier, video samples in ViMUL-Bench are grouped into three categories based on their duration: short \textit{(0-4 mins)}, medium \textit{(4-15 mins)}, and \textit{long (15+ mins)}. Fig. \ref{fig:video_durations} illustrates how different methods perform across these categories. Overall, GPT-4o achieves the highest accuracy on short and medium videos but struggles with long videos in a multilingual setting. In comparison, our proposed baseline, ViMUL, outperforms all LMMs on long videos and surpasses LLaVA-OneVision on both short and medium videos. Most open-source models perform best on short videos, with their accuracy struggling on medium and long videos. The only exception is ViMUL, due to its extensive multilingual training corpus.
Notably, the difference between short and long video performance is comparable in many open-source methods, possibly due to the fewer long videos as compared to short videos in ViMUL-Bench, as shown in Fig. 13 (suppl. material). The culturally curated long videos are harder to find and expensive to manually annotate.

\section{Assessing the need for a Multilingual Video Benchmark}
\label{need-for-multilingual}
To address this question, we conduct three baseline ablations on ViMUL-Bench. \textbf{(1)} \textit{Blind Baseline:} We evaluate LMMs using only the textual QA pairs, without providing the visual input. As shown in Fig. \ref{fig:language_only} (suppl. material), the performance of LLM-only variants drops significantly in the absence of video input. This demonstrates the necessity of incorporating visual input to ensure fair and comprehensive model evaluation. \textbf{(2)} \textit{Image LMM Baseline:} We prompt LLaVA-OneVision and ViMUL with single frames instead of the complete video (32 frames). For rigorous evaluation, we test it on the \textit{First}, \textit{Middle}, \textit{Last}, and \textit{Random} frames. As summarized in Tab.~\ref{tab:frame_experiment} (suppl. material), the performance of LLaVA-OneVision drops by 8.9\% when prompted with a random frame and by 12\% when prompted with the first frame. A similar trend is observed for ViMUL. \textbf{(3)} \textit{Performance on a controlled benchmark:}We conduct an additional experiment using the controlled benchmark CVRR-ES~\cite{khattak2024good} to assess model performance before and after fine-tuning on ViMUL-Instruct. We sample eight spatiotemporal dimensions from CVRR-ES and present the results in Fig. \ref{fig:finetuning_before_after} (suppl. material). Categories such as \textit{Non-Existent Actions with Existent Scene Depictions} and \textit{Interpretation of Social Context} show improvements of 27.68\% and 14.68\%, respectively. Other categories, including \textit{Interpretation of Visual Context} and \textit{Multiple Actions in a Single Video}, also demonstrate consistent performance gains. However, categories like \textit{Understanding Emotional Context} and \textit{Partial Actions} exhibit comparatively lower improvement.

\begin{figure}[!t]
    \centering
    \includegraphics[width=0.5\textwidth]{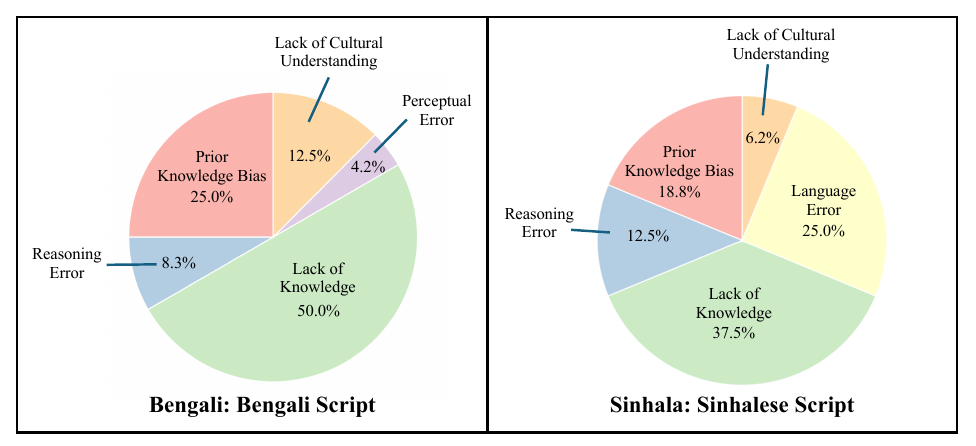}
    \vspace{-1.5em}
    \caption{\textbf{Performance of different question types in ViMUL-Bench}. Overall, the MCQs attain better performance than the open-ended QAs. ViMUL achieves competitive performance compared to existing open-source models.}
    \label{fig:pie_chart}

    \vspace{-1.5em}
\end{figure}

\section{Error analysis of GPT-4o by native speakers}

Fig. \ref{fig:pie_chart} displays an error analysis of GPT-4o’s performanceby native speakers on ViMUL-Bench across Bengali and Sinhalese scripts, revealing distinct patterns of failure. In the Bengali script, the most frequent issues stem from Prior Bias Knowledge and Lack of Knowledge, which are 25\% and 50\%, respectively. While in the Sinhalese script, errors are more often attributed to Lack of Knowledge and Language Error, which are 37.5\% and 25\%, respectively. These disparities highlight the complex interplay between language structure, cultural grounding, and model capabilities. ViMUL-Bench effectively surfaces such weaknesses, emphasizing the need for culturally and linguistically grounded evaluation in multilingual VQA systems.

\section{License for Artifacts}

We use the following datasets and models in our paper. We list all the licenses for each of them below

\begin{itemize}
    \item Base LLMs and the multimodal finetuned models accessed via Hugging Face (e.g., Video-ChatGPT-7B, LLaVA-NeXT-Video-7B,, VideoChat2-Mistral-7B, VILA v1.5 - 7B, LLaVA OneVision - 7B, GPT-4o, GPT-4o-mini, Qwen-3, LLaMA-3.1, Phi-4) are cited with their respective papers, model cards, and URLs. 

    Licenses for multimodal LMMs:
    \begin{itemize}
        \item Video-ChatGPT-7B: CC BY 4.0
        \item LLaVA-NeXT-Video-7B: LLAMA 2 Community License
        \item VideoChat2-Mistral-7B: Apache License 2.0
        \item VILA-v1.5-7B: Apache License 2.0 (code), CC BY-NC-SA 4.0 (weights)
        \item LLaVA-OneVision-7B: LLAMA 2 Community License
        \item Qwen-3: Apache License 2.0
        \item Phi-4: MIT License
        \item GPT-4o: Closed-source (OpenAI Proprietary)
        \item GPT-4o-mini: Closed-source (OpenAI Proprietary)
        \item LLaMA-3.1: LLaMA 3 Community License (restricted commercial use)
    \end{itemize}

    Licenses for the datasets used in our work for the construction of fine-tuning and evaluation benchmark.
    \begin{itemize}
        \item \textbf{CVRR}: CC BY-NC-SA 4.0
        \item \textbf{VideoMME}: Academic research only; commercial use prohibited
        \item \textbf{VCG Diverse}: MIT License
        \item \textbf{MVBench}: MIT License
        \item \textbf{Video-Instruct100K}: CC BY-SA 4.0
        \item \textbf{LLaVA-Video-178K}: Apache License 2.0
        \item \textbf{NeXT-QA}: MIT License
        \item \textbf{PerceptionTest}: CC BY 4.0
        \item \textbf{CLEVRER}: CC0 License
        \item \textbf{ActivityNetQA}: MIT License (non-commercial use only)
    \end{itemize}

    \item The video samples for cultural category was scraped manually from the publicly accessible websites, such as YouTube. The usage of this content is compliant with fair-dealing law for non-commercial academic research. We do not redistribute the original video and text data under commercial licensing. 

    \item Our codebase builds on open-source frameworks such as PyTorch (BSD-style license) and Transformers (Apache 2.0). Our evaluation framework is based on the lmms-eval framework, based on Apache License 2.0.
    
\end{itemize}
The terms of use for each resource are respected, and we provide links and citations in the supplementary material and code repository. We release our work under the license \href{https://creativecommons.org/licenses/by-nc/4.0/deed.en}{CC BY-NC 4.0}.

\section{Documentation of Artifacts}

We list the coverage of domains of the datasets used in our paper:

\subsection{Evaluation Benchmark Domains}

\begin{itemize}
    \item \textbf{CVRR:}
    \begin{itemize}
        \item Non-existent actions with non-existent scene depictions
        \item Non-existent actions with existent scene depictions
        \item Time order understanding
        \item Understanding of emotional context
        \item Interpretation of social context
        \item Unusual and Physically Anonymous activities
        \item Continuity and Object Instance Count
        \item Interpretation of visual context
        \item Partial actions
        \item Fine-grained action understanding
        \item Multiple actions in a single video
    \end{itemize}

    \item \textbf{VideoMME:}
        \begin{itemize}
            \item Domains in dataset:
                \begin{itemize}
                    \item Life Record: Travel, Daily Life, Fashion, Food, Handicraft, Pet \& Animal, Exercise, Multilingual
                    \item Artistic Performance: Acrobatics, Variety Show, Magic Show, Stage Play
                    \item Sports Competition: Other Sports, Athletics, Football, Basketball, Esports
                    \item Film \& Television: News Report, Documentary, Movie \& TV Show, Animation
                    \item Knowledge: Technology, Life Tip, Law, Geography, Astronomy, Finance \& Commerce, Biology \& Medicine, Literature \& Art, Humanity \& History
                \end{itemize}

            \item Domains in various questions:
                \begin{itemize}
                    \item Temporal Perception
                    \item Spatial Perception
                    \item Attribute Perception
                    \item Action Recognition
                    \item Object Recognition
                    \item OCR Problems
                    \item Counting Problems
                    \item Temporal Reasoning
                    \item Spatial Reasoning
                    \item Action Reasoning
                    \item Object Reasoning
                    \item Information Synopsis
                \end{itemize}

        \end{itemize}

    \item \textbf{MVBench:}
        \begin{itemize}
            \item Spatial Understanding
                \begin{itemize}
                    \item Action: What's the man doing?
                    \item Object: What's on the table?
                    \item Position: Is the man on the stage?
                    \item Count: How many chairs?
                    \item Scene: Where's the man?
                    \item Pose: What's the man's pose?
                    \item Attribute: What color is the desk?
                    \item Character: What are the subtitles?
                    \item Cognition: Why is the man singing in the canteen?
                \end{itemize}

            \item Temporal Understanding

                \begin{itemize}
                    \item Action: Action Sequence, Action Antonym, Action Prediction, Unexpected Action, Fine-grained Action
                    \item Object: Object Shuffle, Object Existence, Object Interaction
                    \item Position: Moving Direction, Action Localization
                    \item Count: Action Count, Moving Count
                    \item Scene: Scene Transition
                    \item Pose: Fine-grained Pose
                    \item Attribute: State Change, Moving Attribute
                    \item Character: Character Order
                    \item Cognition: Episodic Reasoning, Egocentric Navigation, Counterfactual Inference
                \end{itemize}

        \end{itemize}

    \item \textbf{VCG-Diverse}

        \begin{itemize}
            \item Video Domains
                \begin{itemize}
                    \item News, Surveillance, Traffic, Automobiles, Sports, Gaming, Cooking, HowTo, Travel, Pets, Education, Science, Entertainment, Music, Comedy, Film, Lifestyle, Activism
                \end{itemize}

            \item Question Types
                \begin{itemize}
                    \item Sequential Understanding: Cooking, How-to, Education
                    \item Predictive Reasoning: Sports, Gaming
                    \item World Knowledge: Science, News
                    \item Causal Reasoning: Surveillance, Activism
                    \item Emotional Reasoning: Entertainment, Film, Comedy
                    \item Analytical Reasoning: Traffic, Automobile
                \end{itemize}
                
        \end{itemize}
\end{itemize}

\subsection{Instruction Fine-Tuning Domains}
\begin{itemize}
    \item \textbf{Video-Instruct100K}: Video Summarization, Description-based QA (spatial, temporal, relational, reasoning), Creative/Generative QA
    \item \textbf{LLaVA-Video-178K}: Academic Sources, YouTube Videos, ActivityNetQA, NeXT-QA, PerceptionTest, LLaVA-Hound
    \item \textbf{NeXT-QA}: Causal Action Reasoning, Temporal Action Reasoning, Object Interaction Understanding, Daily Activity Scenarios
    \item \textbf{PerceptionTest}: Memory, Abstraction, Physics, Semantics
    \item \textbf{CLEVRER}: Temporal Reasoning, Causal Reasoning, Physical Dynamics, Symbolic Event Representation
    \item \textbf{ActivityNetQA}: Long-Term Spatio-Temporal Reasoning, Complex Web Video Understanding, Human-Annotated QA Pairs
\end{itemize}

\section{Model Size and Budget}

We had the following compute budget for our project:

\begin{itemize}
    \item \textbf{Evaluation on ViMUL-Bench (including ablations):}
    \begin{itemize}
        \item Total GPU Hours: 9
        \item GPU Variant: AMD-MI200 (64GB)
    \end{itemize}

    \item \textbf{Finetuning ViMUL on ViMUL-Instruct:}
    \begin{itemize}
        \item Total GPU Hours: 200--240
        \item GPU Variant: AMD-MI200 (64GB)
    \end{itemize}

    \item \textbf{Translation of 130k samples for \textit{cycle consistency} using Qwen-3:}
    \begin{itemize}
        \item Total GPU Hours: 12,000
        \item GPU Variant: AMD-MI200 (64GB)
    \end{itemize}
\end{itemize}

\section{Experimental Setup and Hyperparameters}

\paragraph{Frame Sampling Hyperparameter}
In our experimental setup, we investigated the impact of the number of frames used for video representation. We defined $N$ as the maximum number of frames to be processed per video. While the default setting for ViMUL uses $N = 32$ uniformly sampled frames at 1 FPS, we also conducted ablation studies with single-frame inputs (e.g., using only the first, middle, or random frame). Our results showed that such single-frame configurations led to substantial drops in performance, especially in spatio-temporal and reasoning-heavy tasks. Therefore, $N = 32$ was selected as the optimal configuration, offering a strong balance between computational efficiency and model accuracy.

\paragraph{Evaluation Hyperparameter}
We evaluated the ViMUL-Bench on the following hyperparamters in Tab. \ref{tab:evaluation_hyperparamter}

\begin{table}[h!]
\centering
\begin{tabular}{ll}
    \toprule
    \textbf{Parameter}       & \textbf{Value} \\
    \midrule
    max\_new\_tokens         & 1024           \\
    temperature              & 0              \\
    do\_sample                & False          \\
    num\_beams                & 1              \\
    batch\_size               & 1              \\
    \bottomrule
\end{tabular}
\caption{Evaluation hyperparameters used during inference.}
\label{tab:evaluation_hyperparamter}
\end{table}
\paragraph{Finetuning Hyperparameters}

We fine-tune ViMUL-OneVision using the following key hyperparameters:.

\begin{itemize}
    \item Base LLM: DeepSeek-R1-Distill-Llama-8B
    \item Vision Tower: google/siglip-so400m-patch14-384
    \item Projector Type: mlp2x\_gelu
    \item MM Tunable Parts:  mm\_mlp\_adapter, mm\_language\_model
    \item Learning Rate: 1e-5
    \item Weight Decay: 0.0
    \item Train Batch Size: 2
    \item Eval Batch Size: 1
    \item Gradient Accumulation: 1
    \item Epochs: 1
    \item Warmup Ratio: 0.03
    \item LR Scheduler: Cosine
    \item Precision: bf16
    \item Gradient Checkpointing: True
    \item Max Token Length: 8192
    \item Torch Compile: True (inductor)
    \item Deepspeed Config: Zero3
    \item Save Steps: 1000
    \item Save Total Limit: 1
\end{itemize}

\section{Instruction to Volunteers}
\subsection{Cultural Video Dataset Curation}
For each language, we have identified eight distinct cultural categories. We gave the following instructions to the volunteers to collect the videos:

\begin{itemize}
    \item From your selected language choose 3 to 4 videos from each of the 8 cultural categories
    \item Please avoid choosing videos that contain sensitive personal information such as private addresses, Social Security numbers, or any other confidential data.
    \item Whenever possible, select videos filmed in public places to respect privacy. Ensure that the video content does not infringe on the privacy of any individuals or groups. 
    \item Videos should not depict or disclose any private or sensitive content without the consent of the people involved.
    \item Always ensure that the content of the video respect the privacy of individuals and do not include private or sensitive information.
    \item If there is any doubt about the appropriateness of a video, please consult the project supervisor or team for clarification.
    \item Ensure that the video is public and its license is also public.
\end{itemize}

\subsection{Cultural QA Curation}
After the videos are collected, we asked our volunteers to curate QA pairs of both multiple choice and open-ended QA pairs, for each video in their respective native language. The instructions were given as follows:

\begin{itemize}
    \item Please watch each video carefully. If you feel that the video does not align with the specified category or the native language, you are welcome to replace it with a new YouTube video link. Ensure that the video is public and its license is also public.
    \item After watching the video, you are required to create 3 Multiple Choice Questions (MCQs) and 1 Short Answer Question (SAQ) in English based on the video's content.
    \item Additionally, for each question, you must provide a translated version in the native language along with the answer.
\end{itemize}

\subsection*{Instructions for Writing Multiple Choice Questions (MCQs)}

\begin{itemize}
    \item \textbf{Video-Dependent Questions:} Ensure that each question requires the viewer to watch the video to answer. Avoid questions that could be answered with general knowledge or basic text comprehension. \textit{For example:} "Where is the University of Central Florida?" is incorrect, while "Where is the university shown in the video?" is correct.
    
    \item \textbf{Contextual Understanding:} The question should assess the viewer's understanding of the video's context, including spatial relationships, object interactions, and scene transitions. \textit{For example:} "Where does the car park after crossing the street, as shown in the video?" is a suitable question.
    
    \item \textbf{Avoid Ambiguity:} Ensure that all questions and answer options are clear, unambiguous, and directly linked to the video content. Vague or open-ended questions can confuse models and reduce the benchmark's effectiveness. \textit{For example:} Instead of asking "What happens next?", use "What does the person do immediately after sitting down?"
    
    \item \textbf{Multiple Choices with Plausible Distractors:} Distractors (incorrect options) should be plausible and require watching the video carefully to rule out. Distractors may involve objects or events that appear similar to the correct answer or occur at different times in the video. \textit{For example:} In the question "What does the person eat in the video?", possible choices could be:
    \begin{itemize}
        \item A. A sandwich
        \item B. An apple
        \item C. A book (implausible distractor)
        \item D. A banana (plausible distractor)
    \end{itemize}
    
    \item \textbf{Cultural or Contextual Awareness:} Ensure that questions are culturally sensitive and relevant to the video's context, especially if the video pertains to specific cultural events or actions. This will help avoid bias and make the dataset more generalizable. \textit{For example:} "What festival is being celebrated in the video based on the decorations and activities?"

    \item \textbf{MCQs Format:} The MCQs should be written in the following format:
    \begin{itemize}
        \item \textbf{Question:} <Question>
        \item \textbf{Answer:} Correct\_Answer (A. Option 1, B. Option 2, C. Option 3, D. Option 4)
    \end{itemize}
    
    \item \textbf{Example MCQ:}
    \begin{itemize}
        \item \textbf{Question:} What traditional Japanese garment are the individuals in the Video wearing?
        \item \textbf{Answer:} Kimono (A. Kimono, B. Sari, C. Hanbok, D. Cheongsam)
    \end{itemize}
    
\end{itemize}

\subsection{Instructions for Writing Short Answer Questions (SAQs)}

\begin{itemize}
    \item \textbf{Concise and Precise Questions:} The questions should be brief and to the point, avoiding unnecessary complexity. \textit{For example:} "What color is the car that drives by at the beginning?"
    
    \item \textbf{Culturally Relevant Questions:} If the video depicts culturally specific events or actions, ensure that the question is contextually appropriate for that setting. This ensures relevance and avoids bias. \textit{For example:} "What festival is being celebrated in the video?" (based on visual cues like decorations or attire).
    
    \item \textbf{Answer Length:} Answers should be concise, generally between 1-10 words.
\end{itemize}

\subsection{Verifying Phi-4 Scores}

Volunteers are tasked with verifying the phi-4 scores assigned to model predictions for a set of videos. The verification process involves reviewing the video, the question, the ground truth answer, the model's prediction, and the assigned score. Based on this review, you will determine whether the assigned score is appropriate or if it needs adjustment.

\subsection*{Step-by-Step Instructions}

\begin{enumerate}
    \item \textbf{Watch the Video:} Using the \texttt{mint\_video\_id}, watch the video associated with each QA.
    
    \item \textbf{Review the Question, Ground Truth Answer, and Model Prediction:} Carefully read the question (\texttt{Q}), the ground truth answer (\texttt{A}), and the model's prediction (\texttt{Prediction}) in the provided columns.
    
    \item \textbf{Check the Assigned Score:} Look at the score already assigned in the \texttt{Score} column. The score ranges from 0 to 5, with increments of 0.5.
    
    \item \textbf{Assess the Model's Prediction:} Ask yourself: \textit{Does the model's prediction deserve the score it was given compared to the ground truth answer?}
    
    \item \textbf{Select Your Response:}
    \begin{itemize}
        \item If \textbf{YES}, indicating that the model's prediction is appropriately scored, select \textbf{YES} from the dropdown menu under the \texttt{Do you agree with the score?} column. Leave the \texttt{Your Score} column empty.
        \item If \textbf{NO}, indicating that the model's prediction does not deserve the score assigned, select \textbf{NO} from the dropdown menu under the \texttt{Do you agree with the score?} column. Then, select your preferred score from the dropdown menu in the \texttt{Your Score} column.
    \end{itemize}
\end{enumerate}

\subsection{Verifying Machine Translated Generic Category QA pairs}

Volunteers are tasked with verifying the machine-translated QA pairs for different languages. There are two types of QA pairs: Open-ended and Multiple Choice Questions (MCQs). The goal is to verify the accuracy of the translations for both types of questions and answers. If the translation is correct, the volunteer will mark it as "YES"; if incorrect, they will correct the translation and provide the updated information in the appropriate columns.

\subsection*{Step-by-Step Instructions}

\begin{enumerate}
    \item \textbf{Understand the Columns:}
    The Excel sheet contains the following columns for English (Ground Truth):
    \begin{itemize}
        \item \texttt{English\_Question}
        \item \texttt{English\_MCQ\_Choice\_1}, \texttt{English\_MCQ\_Choice\_2}, \texttt{English\_MCQ\_Choice\_3}, \texttt{English\_MCQ\_Choice\_4}
        \item \texttt{English\_Answer}
    \end{itemize}
    
    The following columns are for the translations:
    \begin{itemize}
        \item \texttt{Translated\_Question}
        \item \texttt{Translated\_MCQ\_Choice\_1}, \texttt{Translated\_MCQ\_Choice\_2}, \texttt{Translated\_MCQ\_Choice\_3}, \texttt{Translated\_MCQ\_Choice\_4}
        \item \texttt{Translated\_Answer}
    \end{itemize}
    
    \item \textbf{Check the Translation Accuracy:} The English text serves as the Ground Truth. For each row, review the translated versions of the question, MCQ choices, and answer.
    
    \item \textbf{Verify the Translation:}
    \begin{itemize}
        \item If the translation is \textbf{correct}, write \texttt{YES} in the \texttt{Is the translation correct?} column.
        \item If the translation is \textbf{incorrect}, write \texttt{NO} in the \texttt{Is the translation correct?} column, and insert the \textbf{correct translation} in the respective columns:
        \begin{itemize}
            \item {Correct\_Translated\_Question}
            \item {Correct\_Translated\_MCQ\_Choice\_1} 
            \item {Correct\_Translated\_MCQ\_Choice\_2} 
            \item {Correct\_Translated\_MCQ\_Choice\_3} 
            \item {Correct\_Translated\_MCQ\_Choice\_4}
            \item {Correct\_Translated\_Answer}
        \end{itemize}
    \end{itemize}
    
    \item \textbf{Handling Open-ended QA Pairs:} 
    \begin{itemize}
        \item For Open-ended QA pairs, the MCQ columns will be empty.
        \item If the question is Open-ended and the translation is incorrect, only fill in the \texttt{Correct\_Translated\_Question} column and leave the MCQ and answer columns empty.
    \end{itemize}
\end{enumerate}

\subsection*{Example}

\textbf{Incorrect Translation Example:} 

\begin{itemize}
    \item \textbf{If the translated question is incorrect:} 
    \begin{itemize}
        \item Write \texttt{NO} in the \texttt{Is the translation correct?} column.
        \item Insert the correct translation in the \texttt{Correct\_Translated\_Question} column.
        \item Leave the other columns (MCQ choices and answer) empty if they are correct.
    \end{itemize}
\end{itemize}

\begin{table*}[h]
    \centering
    \resizebox{\textwidth}{!}{ 
    \begin{tabular}{lllll}
        \toprule
        \textbf{Languages} & \textbf{Country} & \textbf{Script} & \textbf{Family} & \textbf{Specification} \\
        \midrule
        Arabic   & UAE, Saudi, Egypt  & Arabic   & Afro-Asiatic  & High \\
        Bengali  & Bangladesh, India  & Bengali  & Indo-European & High \\
        Chinese  & China              & Chinese  & Sino-Tibetan  & High \\
        French   & France             & Latin    & Indo-European & High \\
        German   & Germany            & Latin    & Indo-European & High \\
        Hindi    & India              & Devanagari & Indo-European & High \\
        Japanese & Japan              & Kanji/Kana & Japonic      & High \\
        Russian  & Russia             & Cyrillic  & Indo-European & High \\
        Sinhala  & Sri Lanka          & Sinhalese & Indo-European & Low \\
        Spanish  & Spain              & Latin    & Indo-European & High \\
        Swedish  & Sweden             & Latin    & Indo-European & High \\
        Tamil    & India              & Tamil    & Dravidian     & Low \\
        Urdu     & Pakistan           & Arabic   & Indo-European & Low \\
        \bottomrule
    \end{tabular}
    }
    \caption{Language Classification by Country, Script, Family, and Specification}
    \label{tab:language_classification}
\end{table*}